\documentclass{iopjournal}

\usepackage{algorithm}
\usepackage{algpseudocode}

\usepackage{array}
\usepackage{tabularx}
\usepackage{booktabs}

\usepackage{xcolor}
\usepackage[most]{tcolorbox}
\usepackage{mdframed}
\usepackage{svg}
\usepackage[numbers,sort&compress]{natbib}

\usepackage[T1]{fontenc}
\usepackage{lmodern}
\usepackage{microtype}
\usepackage{markdown}
\usepackage[skip=0.8\baselineskip]{parskip}
\usepackage[english]{babel} 
\usepackage{ragged2e}
\justifying

\usepackage{caption}
\usepackage[subpreambles=true]{standalone}                                                                                                         
\usepackage{graphicx}   
\graphicspath{{task_gausfit/}{task_calovq/}}

\definecolor{lightgraybox}{gray}{0.96}

\begin{document}

\articletype{} 

\title{\textbf{SciFi}: A Safe, Lightweight, User-Friendly, and Fully Autonomous Agentic AI Workflow for Scientific Applications}

\author{Qibin Liu$^{1*}$, Julia Gonski$^1$}

\affil{$^1$SLAC National Accelerator Laboratory, 2575 Sand Hill Rd, Menlo Park CA 94025 USA }

\affil{$^*$Author to whom any correspondence should be addressed.}

\email{qibin@slac.stanford.edu}

\begin{abstract}
Recent advances in agentic AI have enabled increasingly autonomous workflows, but existing systems still face substantial challenges in achieving reliable deployment in real-world scientific research. 
In this work, we present a safe, lightweight, and user-friendly agentic framework for the autonomous execution of well-defined scientific tasks. 
The framework combines an isolated execution environment, a three-layer agent loop, and a self-assessing do-until mechanism to ensure safe and reliable operation while effectively leveraging large language models of varying capability levels. 
By focusing on structured tasks with clearly defined context and stopping criteria, the framework supports end-to-end automation with minimal human intervention, enabling researchers to offload routine workloads and devote more effort to creative activities and open-ended scientific inquiry.
\end{abstract}

\section{Introduction}

An agentic system is an AI system that pursues a user-defined goal through an iterative closed loop of planning, action, observation, and revision, rather than producing only a single-pass response~\cite{wang2023survey}. 
In the context of LLMs, key ingredients typically include task decomposition, interaction with external tools or execution environments, and feedback-driven refinement based on intermediate results~\cite{yao2023react, shinn2023reflexion}. 
These properties enable goal-directed problem solving under verification, which is particularly relevant for complex scientific and engineering workloads where success often requires multi-step reasoning, tool use, error recovery, and repeated improvement driven by quantitative feedback.

Recent advances in agentic AI systems have enabled increasingly autonomous workflows across many domains, such as self-driving cars~\cite{7490340} and financial trading~\cite{7407387}. The introduction of agents to automate and accelerate scientific tasks is just beginning, but early results are already promising~\cite{gridach2025agenticaiscientificdiscovery, schwartz2026resummationcparametersudakovshoulder,laurent2024labbenchmeasuringcapabilitieslanguage,kim2024mdagentsadaptivecollaborationllms,kumbhar2025hypothesisgenerationmaterialsdiscovery}. 
However, many current agentic systems either remain research prototypes without widely reusable tooling, or they target general-purpose tasks with complex and partially opaque architectures that do not fully address the requirements of reliably unattended execution in scientific contexts. 

A large fraction of scientific work is time-consuming not because its goals are unclear, but because its execution is highly customized. Tasks such as intent-driven data visualization, domain-informed optimization, and instrumentation workflows including firmware or data-acquisition (DAQ) debugging are often easy to specify at a high level and straightforward to verify after the fact, yet difficult to operational convert into reusable executable procedures. They typically require iterative trial and error and close interaction with specialized tools, shared computing infrastructure, or laboratory environments. We refer to such problems as \emph{closed-loop scientific tasks}: domain-specific workflows with clear objectives, explicit constraints, and verifiable stopping criteria. 
Existing agentic systems are often ill-matched to this regime, as they either depend on brittle one-shot generation of detailed execution steps, or require frequent human supervision which limits their practical value as time-saving research infrastructure.

In this work, we present \textbf{SciFi} as a safe, lightweight, and user-friendly agentic framework designed specifically for closed-loop scientific tasks. 
The framework is built around three closely connected principles: safety, usability, and extensibility. First, \textit{safety} is achieved through an isolated execution environment that supports unattended operation, prevents unintended side effects on shared computing systems, and improves reproducibility by controlling dependencies and runtime state. Second, \textit{usability} is enabled by a closed-loop autonomous pipeline, defined by a three-layer agent loop coupled to a self-assessing "do-until" mechanism, allowing the system to iteratively refine its outputs, detect failures, and stop only when explicit task criteria are satisfied. This design reduces the need for detailed user supervision and allows for easy exchanging of backbone large language models (LLMs) with different capability levels, rather than relying solely on the strongest models. 
Third, \textit{extensibility} is enabled by a minimalist, LLM-native framework design that combines descriptive system prompts with fixed deterministic components, allowing for natural-language task bootstrapping, making the system easier to adapt across tasks through recursive self-improvement.

Related prior work has explored several key directions: (i) interleaving reasoning and action in tool-augmented agents, (ii) learning to invoke external tools via APIs and programmatic interfaces, (iii) modular architectures that integrate LLMs with external reasoning components, (iv) multi-agent orchestration through conversational protocols, (v) iterative improvement via reflection and textual memory, and (vi) LLM-assisted conversion and reproduction of workflow~\cite{yao2023react,schick2023toolformer,gao2023pal,karpas2022mrkl,wu2024autogen, shinn2023reflexion,masera2025snakemakerseamlesslytransformingadhoc}. The focus of SciFi differs in explicitly targeting the closed-loop scientific regime, where task boundaries and stopping conditions are well-defined but execution is highly customized and reliability constraints are stringent.
While closed-loop tasks are the design driver, in order to fully describe of SciFi's capability to address the full suite of tasks in scientific research, its performance on open-ended creative tasks is also of interest.

We thus present the application of SciFi to a suite of representative scientific tasks: basic daily research tasks, data analysis and visualization, generation of machine learning (ML) methods, hardware-related code optimization, and open-ended challenges.
For each experiment we report the agent's iteration count to completion, wall-clock cost, human and external intervention, and a brief iterations analysis to understand the robustness of the workflow.
\textbf{SciFi provides robust and verifiably correct results across all scientific tasks considered here}, with a yet-uncalculable decrease in time-to-result compared to human researchers. 
This work lays foundational steps toward the safe, efficient use of agentic systems coupled with appropriate human-in-the-loop activity to accelerate scientific discovery.

\section{Agentic AI System and Design Principles}

As the SciFi framework was specifically designed for scientific applications, it thus has several three key features that distinguish it from related works as highlighted earlier.
Technical details on the incorporation of each principle are provided below. 

\paragraph{Safety and resource isolation.}
The first design principle is to enable safe unattended execution. To reduce human intervention without risking unintended side effects on shared infrastructure, the entire workflow (including the agent processes and any spawned tools) executes inside an isolated container runtime (e.g., \texttt{Apptainer}~\cite{singularity_zenodo_2021,kurtzer2017singularity}). Resource access is governed by explicit, task-scoped rules, including GPU usage, network access, read-only data mounts, and read/write scratch storage.

Isolation relies on Linux-kernel-supported mechanisms and on container filesystem primitives. In particular, \texttt{Apptainer} uses an immutable container image format whose filesystem is stored as a \texttt{SquashFS} object, with overlay mechanisms (including \texttt{fuse-overlayfs} when required) used to provide controlled write layers. Resource mappings are declared in the task description and resolved \emph{before} the agent loop begins, following a default-deny logic: if a resource is not explicitly specified, it is not mapped from the host. As a result, the accessible resources are task-specific, explicit, auditable, and bounded by construction.

\paragraph{Usability across models and levels of intervention.}
A second design goal is high usability. The system is designed to operate with a wide range of backbone models (including both open-weight and closed commercial models) and to remain functional under different levels of human intervention and different degrees of completeness in the initial task description. The aim is to provide a vendor-unlocked, open-box solution that bridges scientific workloads and modern agentic capabilities.

To achieve this, the system combines a self-assessing task structure with a three-layer agent loop (bounded in practice by hard limits for budget control), together with a model-gateway interface (e.g., \texttt{LiteLLM}~\cite{berriai_litellm}) and a model-ranking mechanism. The ranking mechanism assigns models of different capability and cost to different sub-tasks, reducing overall cost while improving availability via interchangeable load balancing among models of the same rank.

Within this design, the agent loop searches for solution trajectories that connect a fixed starting point (e.g., input data, initial conditions, or an existing code base) to a fixed end point (e.g., a verifiable output or a quantitative metric), while keeping intermediate steps flexible. We view this as an iterative refinement process driven by verification signals; empirically, we expect the approach to be most effective when the task provides complete causal context and strong, explicit stopping criteria.

\paragraph{Extensibility and LLM-native implementation.}
Finally, the system is \emph{LLM-native}. Only the minimal set of functions that must remain deterministic and trusted are implemented as fixed code, including safety control, container management, the model-gateway interface, tool invocation, and parsing/validation of the initial task description. All other behavior is defined through system prompts or modular \emph{skills}, i.e., reusable predefined context-and-prompt units. This design reduces fixed code, increases flexibility, and makes the system more transparent to both users and the LLM. 

On top of this foundation, we implement LLM-based utilities such as a \emph{task maker} and a \emph{skill maker} that translate user needs into structured task descriptions and reusable skills. We also provide an \emph{ask} tool to support early-stage usage and serve as a natural-language user manual.
As a consequence, the system can be initialized from purely natural-language input, reducing the effort required to draft structured tasks and configure the execution environment. At the same time, the system preserves full openness: task descriptions can still be written by hand, and the framework remains open-box for review, debugging, and extension.

\section{Implementation}
\label{sec:impl}

\begin{figure}[htbp]
    \centering
    \includegraphics[width=0.9\linewidth]{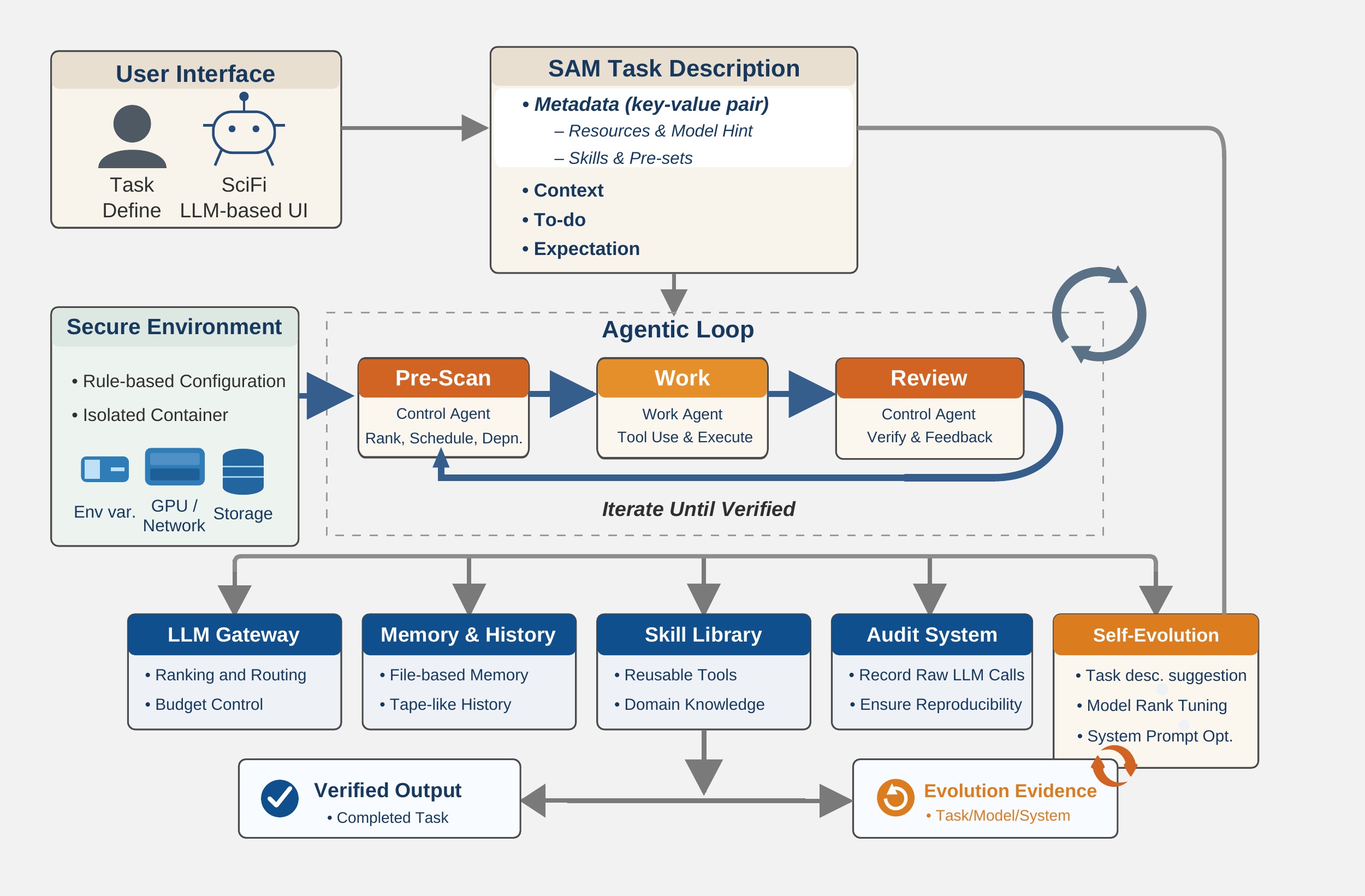}
    \caption{Architecture of the SciFi agentic system, indicating the safety ensured by secure environmental operation, usability provided by a three-agent loop structure working under concrete "do-until" conditions with easy LLM swapping, and extensibility via a natural language interface.}
    \label{fig:agent_system}
\end{figure}

The implementation of the SciFi system is illustrated in the Figure~\ref{fig:agent_system} with each part described in detail in this section. Full implementation details can be found in the codes~\footnote{https://github.com/qibin2020/scifi}.

\subsection{SAM Task Description} 
The basic building block of a workload in this system is defined as a \emph{Self-Assessed Module} (SAM). Each SAM describes a task through three parts: \emph{context}, \emph{to-do}, and \emph{expectation}, of which the latter two are mandatory. In this way, a SAM defines the smallest verifiable problem, with a clear initial condition, a required action, and an explicit end point. The agentic system solves each SAM through an iterative agentic loop until the expectation is matched. A SAM is written in plain Markdown format and can be recursive: if a task is decomposed into sub-tasks, each sub-task should follow the same structure, thereby defining the smallest execution--verification element in a uniform way. The system is designed to ensure that all SAMs satisfy their corresponding expectations before the overall task is claimed as accomplished.

The SAM format is deterministically parsed, so that the expectation is always explicitly extracted from the task description and cannot be fabricated by the agent during execution. An example of a task description written in the SAM format is shown in Sec.~\ref{sec:basic_tasks}. Sub-task decomposition is optional, since the agentic system itself has the ability to decompose tasks in natural language. However, such decomposition is not deterministic and is not strictly verified against predefined criteria. In practice, it is often helpful to introduce checkpoints by explicitly defining additional sub-task SAMs along critical problem-solving paths. At the top level, at least one SAM is required to define the task and kick off the agentic loop. All three parts of a SAM are parsed by the agent in natural language and jointly define the iteration process.

In addition, a SAM may include an optional metadata field that specifies the resource requirements of the task, such as wall-time limits, job system permission, GPU usage, storage permissions, and skill selection. This metadata is used only to request specific resources; by default, no read or write access is allowed outside the task folder. The metadata can also provide hints about task difficulty and preferred model variants, helping the system choose a more suitable backbone model and thereby accelerate convergence. Importantly, all metadata are expressed in key--value format and are parsed and processed deterministically without LLM involvement, which improves safety and robustness. During execution, the SAM itself is treated as read-only, and the agent is never allowed to modify the task description, especially the expectation, thus preserving the robustness of the closed-loop process.

\subsection{Agentic Loop} 
Each SAM is executed through an infinite three-layer agentic loop composed of \emph{pre-scan} $\rightarrow$ \emph{work} $\rightarrow$ \emph{review}. The pre-scan agent first reads the SAM task description and determines the sub-task (SAM) structure, execution order, dependencies, matched LLM rank, and any remaining context from the previous iteration. The work agent is then invoked to carry out the \emph{to-do} until the \emph{expectation} is satisfied. Since the system operates in a fully isolated execution environment, tool use by the work agent can be automatically approved. After the work agent claims that the task or sub-task is complete, an independent review agent is invoked to check the expectation and verify that the current SAM has indeed been satisfied.

Any failure, whether due to a false positive or to a time cutoff in the work stage, triggers a new iteration of the same pre-scan--work--review loop. At the end of each failed iteration, the review agent writes the key history and suggested fixes into task-level memory. This information is then read by the pre-scan agent in the next iteration, allowing it to address dependency-related issues and pass more informed context to the work agent. A SAM is considered complete only when all of its sub-SAMs are completed and its own expectation is verified to be met.

\subsection{Memory, History and Audit Systems}
The agent is stateless by design, meaning that unless specific context is provided, it has no knowledge of previous iterations. To maintain continuity across iterations and enable continuous learning, the system introduces both a memory mechanism and a history mechanism. Both are implemented as text-file-based components to minimize overhead.

The memory system is designed as a read--update interface that allows different agents to exchange information and preserve the most important task-related knowledge, such as recurring failure patterns, previously explored paths, and useful intermediate findings. Three levels of memory are implemented. The first is \emph{task-level memory}, which is shared by agents within the same task run; it is read by the pre-review agent and updated by the review agent. The second is \emph{task-group memory}, which is shared across different runs of the same task and therefore supports multiple re-runs of complex tasks that may require an additional outer loop before convergence. This memory is updated by the final task-review agent and read by the pre-scan agent. The third is \emph{global memory}, which is generated from the final task review and summarizes the experience accumulated from individual task executions. In particular, the final task review produces both task-specific suggestions and global suggestions, focusing on possible improvements to SAM design and system design. The resulting global memory is intended for later system self-evolution, where statistics collected over many tasks can be analyzed to propose system-level optimizations, such as improvements to the system prompt.

The history system is similar in purpose but differs in structure: it is treated as a tape and is strictly append-only. It records the status of each iteration and the key decisions made by the agents, such as pass/fail judgments and major changes in search direction. In the current design, the history is mainly read by the review agent when a task or sub-task fails, where the recorded trajectory can help generate a more informed fix. The history is also consumed by the system self-evolution stage as a data source for broader system optimization.

In addition, a standalone audit system is implemented transparently at the interface to the LLM. Every LLM call is recorded for auditing and debugging purposes, providing a single source of truth from which a task run can be reproduced exactly. Over longer timescales, this recording system also provides a possible path toward constructing scientific-workload-oriented fine-tuning data for open-weight backbone models, thereby enabling continuous improvement of the overall system through reinforcement-learning-based post-training methods, e.g.\ Group Relative Policy Optimization (GRPO) \cite{shao2024deepseekmath}.

\subsection{Skill Library for Domain Knowledge} 
\label{sec:skill}
Besides memory and history, which can be shared across different tasks to provide common context and accumulated experience for faster convergence of the agentic loop, the \emph{skill} mechanism~\cite{wang2024voyager,jiang2026agenticskills} provides a reusable way to document successful paths toward specific targets. A skill is a block of contextual text that describes either a verified solution path to a particular target or domain-specific expertise and operational experience that would be difficult, inefficient, or impossible for the agent to discover reliably in a standalone run under limited time. Typical examples include scientific software setup, which often requires special configurations, and job-submission workflows, which depend on environment-specific knowledge that is not, and in some cases should not be, fully exposed to the agent.

In the experiments described in Section~\ref{sec:agentic_workflow}, the skills are found to significantly accelerate the early iterations of the agent loop by helping the agent resolve common technical issues, such as locating the correct environment, and thereby preserving more budget for the actual problem-solving stage. At the same time, comparison tests show that even without the predefined skills, the agent can still finish the task through a more complicated path, which further demonstrates the robustness of the default configuration.

The skill information is documented in the system design notes and is readable by the LLM-based user interfaces, such as the \emph{task maker} and \emph{ask} tools. As a result, appropriate skills can be suggested automatically or inserted later during task refinement, without requiring the user to have complete prior knowledge of the available skill set. In addition, a \emph{skill maker} interface is provided to summarize task memory and history and convert them into new reusable skills over long-term usage, enabling cumulative user-driven improvement of the system.

\subsection{LLM Gateway and Model Ranking System}
The underlying capability of the system is provided by the backbone LLMs, which are interfaced through an LLM-gateway design. This gateway is built on top of \texttt{LiteLLM} and unifies models from different providers, as well as locally hosted models, behind a single access API. Users can supply API keys from different cloud providers or connect locally deployed open-weight models through the gateway. The gateway supports load balancing and fallback across providers, improving availability when a particular service is rate-limited or interrupted. 

On top of this gateway, a \textit{ranking system} is introduced to provide model selection and routing based on capability. The rank of a model is determined mainly by its reliability in tool calling and its reasoning ability for solving complex tasks or subtasks, and the system matches each task to an appropriate model accordingly. The model ranking is tunable, and across repeated runs the self-evolution stage can further adjust these rankings based on accumulated task history and memory. The ranking system also provides simple budget control: when a model exhausts its budget, it is disabled and the system switches to another model in the same rank. More advanced budget management can be handled by the underlying \texttt{LiteLLM}.

Because the agents in the loop play different roles, the system distinguishes between \emph{control agents} and \emph{work agents}. The pre-scan and review agents are treated as control agents, which typically require stronger reasoning ability and, in some cases, optional tool-use capability. By contrast, the work agent is responsible for carrying out the task itself, and therefore requires strong and reliable tool-calling ability; it also usually consumes most of the runtime and token budget, while not always requiring the strongest reasoning model. This separation allows different model types and ranks to be assigned to different agent roles, thereby reducing cost and improving overall availability. The selection of ranked models is handled jointly by the pre-scan agent and deterministic system logic. In particular, the rank of the work agent is determined dynamically during execution by the pre-scan agent and can be revised by the review agent if the current model is found to be underpowered or unnecessarily overpowered for the task.

\subsection{Secure Environment}
The system isolation layer is implemented using Linux container technology. We adopt \texttt{Apptainer}~\cite{kurtzer2017singularity}, which is widely used in scientific computing and supports unprivileged execution on shared systems where users typically do not have root access. This makes it a practical alternative to the more widely used Docker-based container ecosystem in industry~\cite{docker}.

During runtime, read/write permissions and resource access are controlled through explicit bind mounts. By default, only the task folder is accessible to the agent. This design reduces the risk of arbitrary reads and limits potentially destructive write operations over a wider part of the host system. Network access is enabled by default, since the agent may need it for operations such as cloning repositories or checking updated information, although it can also be disabled when a fully isolated execution mode is required. 

\subsection{LLM-based User Interface}
The LLM-native design of the system further simplifies the LLM-based user interface (UI). Since the system design is documented in a structured form, the user interface can parse the system itself and suggest appropriate usage patterns. This outer LLM-based interface operates on top of an underlying deterministic interface, \texttt{SciF}, which actually drives the agent workflow through shell scripting (from which the name SciFi is adotped). Several additional LLM-based tools are also provided, including \texttt{task\_maker}, \texttt{skill\_maker}, and \texttt{ask}, as discussed above. Together, these tools improve usability, provide an additional form of user manual, and enable users to learn the system through natural-language interaction. The typical call signature of SciFi is listed in Fig.~\ref{fig:scifi_cmd}.

\begin{figure}[htbp]
\centering

\begin{tcolorbox}[enhanced, sharp corners, boxrule=0.5pt,
    colframe=framegray, colback=white,
    width=6in,
    left=6pt, right=6pt, top=5pt, bottom=5pt,
    title={\sffamily\bfseries SciFi User Input Examples},
    coltitle=white, colbacktitle=titlebg, fonttitle=\normalsize]

{\scriptsize\sffamily
\textcolor{scifi}{\texttt{SciFi}} = natural language input (LLM routes to command)\quad
\textcolor{scif}{\texttt{SciF}} = deterministic dispatcher
}

\vspace{4pt}

{\scriptsize\sffamily
\begin{tabular}{@{}cl@{\quad}l@{\quad}l@{}}
\toprule
& \textbf{\textcolor{scifi}{SciFi} input} & \textbf{\textcolor{scif}{SciF} command} & \textbf{Effect} \\
\midrule
\clife & \texttt{bootstrap the system}  & \texttt{BOOTSTRAP} & Build containers, start gateway, smoke test \\
\clife & \texttt{start}  & \texttt{START} & Start LLM gateway \\
\cwork & \texttt{run real\_FW\_void}  & \texttt{RUN <task>} & Execute task in container (local or SLURM auto-routed) \\
\cwork & \texttt{train MNIST on 1 GPU}  & \texttt{MAKE --desc "..."} & Create task from NL description (interactive maker) \\
\cwork & \texttt{help me debug this build}  & \texttt{ASK} & Multi-turn LLM-based tools (debug, explore, code) \\
\cwork & \texttt{what's running?}  & \texttt{STATUS} / \texttt{JOBS} & Show gateway status, active tasks, SLURM jobs \\
\cevo & \texttt{suggest improvements}  & \texttt{EVOLVE suggest} & Analyze memory and history, write improvement proposals \\
\cevo & \texttt{tune model ranks}  & \texttt{EVOLVE model} & Adjust model ranking based on memory and history \\
\cevo & \texttt{run maintenance}  & \texttt{MAINTAIN} & suggest + model rank + files archiving \\
\clife & \texttt{stop}  & \texttt{END} & Stop LLM gateway \\
\bottomrule
\end{tabular}
}

\vspace{3pt}
{\scriptsize\sffamily\itshape
\clife\,System\quad \cwork\,Task\quad \cevo\,Evolution\quad
--- SciFi enforces command safety: destructive ops require manual confirmation.}

\end{tcolorbox}

\caption{Example SciFi interface input. General natural language is accepted to drive the system.}                                                                                     
\label{fig:scifi_cmd} 
\end{figure}

\section{Experiments}
\label{sec:agentic_workflow}

To benchmark the fully autonomous agentic workflow for closed-ended scientific workloads, we conduct four experiments using three types of initial task definitions, which serve as the sole input to the system: Simple natural-language descriptions (\emph{NL}); Structured task descriptions obtained by processing the \emph{NL} input with the \texttt{task\_maker}  (see Sec.~\ref{sec:impl}) (\emph{ST}); and Exhaustive (\emph{EX}), expert-level task specifications provided by human experts or external LLMs with advanced domain knowledge (e.g. commercial models such as Anthropic's \texttt{Claude Opus} that are state-of-the-art at the time of writing), including explicit step-by-step execution guidance. For each configuration, we record the complete iteration history of the autonomous agent, together with the final task performance, iteration cost, and total wall time for comparison. 
To maintain a vendor-unlocked tool, our study primarily focuses on results obtained using the open-weight model \texttt{Gemma4} from DeepMind as both the control and work model. 
While the choice of tasks for each experiment focus here on the field of high energy physics (HEP), the general nature of each task makes this work highly extrapolatable to other scientific fields.

\subsection{Basic Scientific-oriented Tasks}
\label{sec:basic_tasks}

\begin{figure}[htbp]
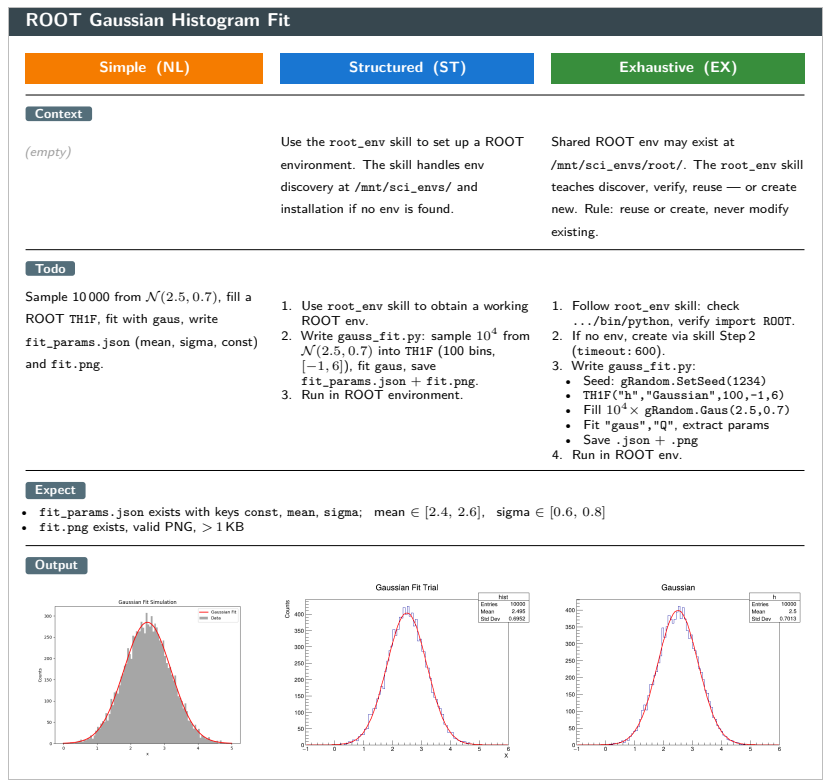

\centering   
\includestandalone[width=0.7\textwidth]{task_gausfit/description}
\caption{Descriptions of the Gaussian fitting task for Experiment 1 at the Simple, Structured, and Exhaustive levels. The output from the workflow is shown on the bottom, where SciFi was successful for all attempts. The Simple description does not strictly enforce ROOT usage in the \texttt{Expect}, so the agent picks the more conventional \texttt{matplotlib} tool.}  
\label{fig:task_gausfit} 
\end{figure}

The first experiment considered covers a broad range of daily work in scientific use case that can be described in a few sentences. These include configuring specific scientific software (for example ROOT~\cite{root_software}, a common tool in HEP), simple data analysis and visualization, and applying/reproducing well-established algorithms to common datasets. One specific example of such a task is the fitting of a Gaussian to a histogram of data using ROOT. 
The task described in \emph{NL}, \emph{ST} and \emph{EX} style is shown in Fig.~\ref{fig:task_gausfit} together with the output from each workflow.

\begin{figure}
    \centering
    \includegraphics[width=1\linewidth]{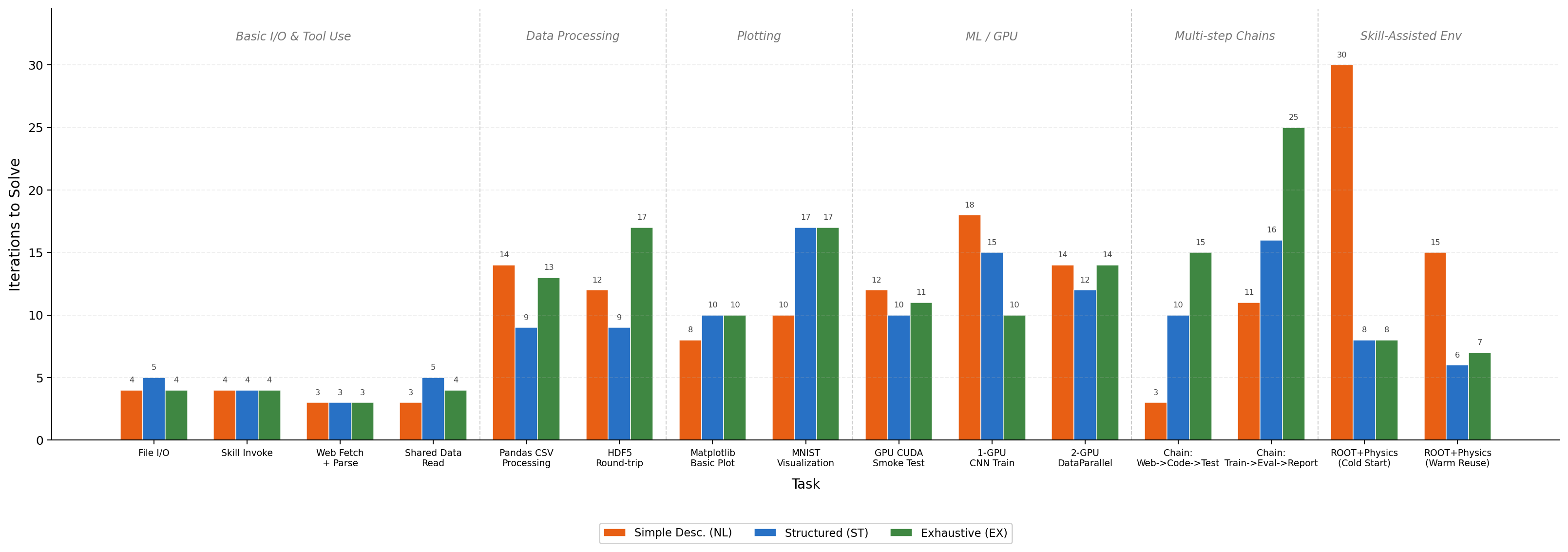}
    \caption{Iteration counts for the agentic loop solving the basic experiment tasks ranging from file I/O, data processing and plotting to ML/GPU related, multi-step and complex environment setup tasks. Results are broken down into Simple, Structured, and Exhaustive task description levels. }
    \label{fig:basic_tasks_iter}
\end{figure}

\begin{figure}
    \centering
    \includegraphics[width=1\linewidth]{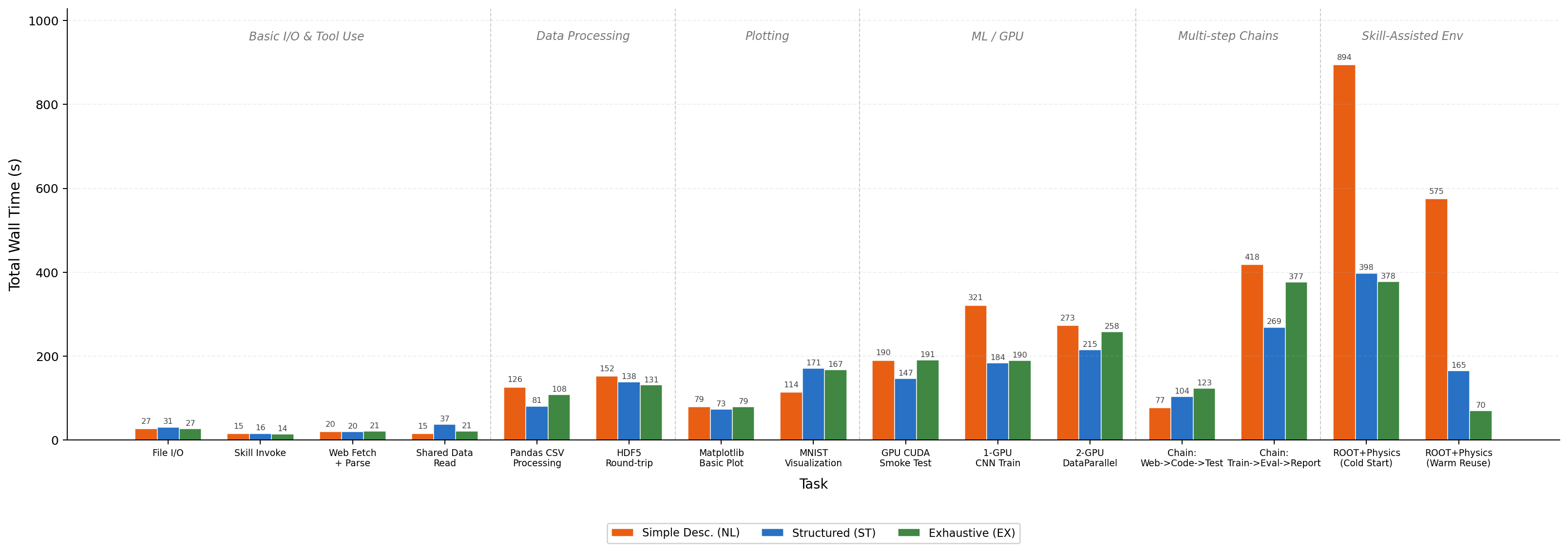}
    \caption{Total walltime including execution time for the agentic loop solving the basic experiment tasks ranging from file I/O, data processing and plotting to ML/GPU related, multi-step and complex environment setup tasks. Results are broken down into Simple, Structured, and Exhaustive task description levels. }
    \label{fig:basic_tasks_wall}
\end{figure}

15 task families across six categories are tested: basic file I/O and tool use (file operations, skill     
  invocation, web fetching, shared data reading), data processing (pandas CSV aggregation, HDF5 round-trip), plotting (matplotlib sine wave, MNIST digit grid), ML/GPU (PyTorch CUDA smoke test, single-GPU and multi-GPU CNN training), multi-step chains (web-to-code-to-test, train-to-evaluate-to-report), and skill-assisted environment setup (cold-start ROOT installation and warm-start reuse via shared environments). 
This set of tasks serves both as a benchmark of SciFi's fundamental capabilities and as a unit test to validate the system. All tests are successfully completed. 
Figures ~\ref{fig:basic_tasks_iter} and \ref{fig:basic_tasks_wall} summarize the results as the number of iterations required to solve each task and the total wall time including execution time, respectively. All tasks are performed in a fully autonomous manner, with only different styles of task descriptions mentioned above to initiate the workflow.

Across Experiment 1 tasks, simple natural-language inputs (NL) solves a large fraction of basic tasks with the fewest iterations and shortest wall time, frequently outperforming more detailed task description such as \emph{ST} and \emph{EX}. This indicates that when the objective and stopping criteria are clear, the closed-loop agent can refine intermediate steps effectively during execution for the basic tasks; excessive up-front instruction may instead constrain the search and increase cost. More structured and detailed descriptions become advantageous mainly for complex, multi-step tasks as described in the subsequent experiments, where explicit planning and intermediate checkpoints reduce exploration.

In practice, the \emph{ST} from \texttt{task\_maker} is most useful as a user interface for suggesting correct system usage, such as recommending relevant skills or metadata controls, rather than as a universally effective decomposition engine. This could be found in the environment installation and ROOT-related tasks where including the proper skills, added by \texttt{task\_maker} when making the \emph{ST} description, does speed up the workflow.
These results suggest a simple usage rule: use \emph{NL} for straightforward tasks, and use structured or expert-informed exhaustive description only for for complex tasks. \emph{NL} is often sufficient for the agent to discover a valid execution path on its own.

\subsection{Closed-Loop Task: Full-Pipeline Reproduction of Published Results}

In the second experiment, we test the system on a common closed-loop task in scientific research, the end-to-end reproduction of paper result. This usually serves as an initial step in a  new research effort. 
It can often be highly time-consuming, but is also well-defined and fully closed-loop with a clear description of the task and expectations, making it a good candidate for agentic assistance.  

The specific example chosen for Experiment 2 is a recent result on the use of generative ML for fast simulation of calorimeters in collider experiments~\cite{liu2024calovqvectorquantizedtwostagegenerative}.
The paper documents a vectorized autoencoder for the calorimeter response tokenization and a transformer-based token generation to create high-fidelity instances of simulated calorimeter response to incident particles. 
The source code for the model architecture and evaluation, along with the test dataset, are fully open sourced and available online.
The target of the task is to fully autonomously set up the environment, get to use of the SLURM job system (with \emph{slurm} skill as mentioned in Sec.~\ref{sec:skill}), run the inference, and then perform the plotting. It tests the ability of the system to drive multi-step, multi-agent workflows for well-defined tasks, aiming to reduce human intervention in fully closed-loop tasks in a fully autonomous way. 

\begin{figure}[!htbp]
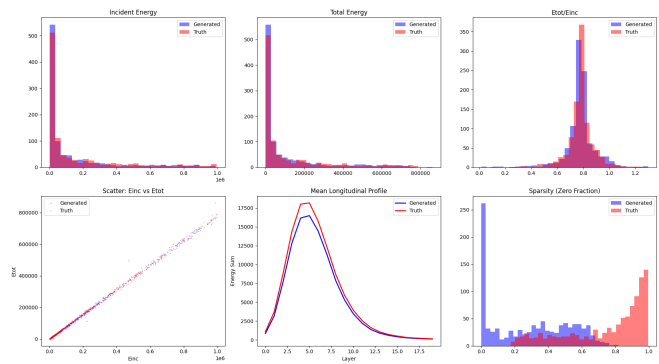

\centering 
\includestandalone[width=0.7\textwidth]{task_calovq/description}
\caption{
Descriptions of the calorimeter simulation task for Experiment 2 at the Simple, Structured, and Exhaustive levels. The output from the workflow is shown on the bottom, showing SciFi success after about 15 minutes.} 
\label{fig:task_calovq} 
\end{figure}

The input and output of the workflow for Experiment 2 is shown in Fig.~\ref{fig:task_calovq}. 
The task description serves as the only condition to start the workflow; the code setup, inference, and plotting are all driven by the SciFi toolkit.
The system operates fully autonomously and successfully completes the task after 69 iterations and about 15 minutes. The progression across different iterations of this run is summarized in Fig.~\ref{fig:history_calovq}. The results demonstrate that the agentic workflow can identify and resolve unexpected issues, such as environment mismatches and unfamiliar computing setups, without user intervention. 
The agent's ability to gain experience solving common issues and achieving common targets during the process can be distilled into reusable skills or memory, enabling more efficient resolution in future runs. This highlights a key design principle of the system: under-specified aspects can be explored autonomously, and increased prior experience and distilled knowledge leads to faster convergence.
This offers a significant reduction of the human scientific effort for well-defined, closed-loop tasks.

\begin{figure}[!htbp]
\centering

\begin{tcolorbox}[enhanced, sharp corners, boxrule=0.5pt,
    colframe=framegray, colback=white,
    width=6in,
    left=8pt, right=8pt, top=6pt, bottom=6pt,
    title={\sffamily\bfseries Calo-VQ Paper Reproduction (Inference Only) --- Iteration History},
    coltitle=white, colbacktitle=titlebg, fonttitle=\normalsize]

{\footnotesize\sffamily
\begin{tabular}{@{}llllll@{}}
\textbf{Work model} & gemma4 &
\textbf{Review model} & gemma4 &
\textbf{Skills} & common\_env, NERSC\_slurm \\
\textbf{Total loops} & 3 (2 exhausted, 1 pass) &
\textbf{Total iters} & 69 &
\textbf{Wall time} & 15\,min\,40\,s \\
\end{tabular}
}

\rowsep

\tsec{Loop\,0\enspace---\enspace 28 iters, 7\,min \hfill \pexh}

{\scriptsize\sffamily
\begin{tabular}{@{}r@{\enspace}l@{\enspace}p{4.6in}@{}}
1--3  & Env        & Found \texttt{/mnt/sci\_envs/calovq}, verified torch + h5py + PL imports \\
4--8  & Clone      & \texttt{git clone}, \texttt{sed} fix for \texttt{running\_sanity\_check}, wrote gen script \\
9--10 & SLURM      & sbatch \pfail: GLIBC\,2.38 required, container has 2.34 \\
11--14 & Download  & 3 curl attempts for 1.3\,GB Zenodo file, all timed out (30\,s default) \\
15--22 & Generate  & Ran locally. Hit git-lfs pointer issue; installed git-lfs, \texttt{git lfs pull} OK \\
23--28 & Generate  & Fixed \texttt{weights\_only} error. Generation running, hit MAX\_ITERATIONS \\
\end{tabular}
}

\rowsep

\tsec{Loop\,1\enspace---\enspace 32 iters, 5.5\,min \hfill \pexh}

{\scriptsize\sffamily
\begin{tabular}{@{}r@{\enspace}l@{\enspace}p{4.6in}@{}}
1--5  & Verify     & \texttt{gen\_ds2.h5} (1000 events) + \texttt{dataset\_2\_2.hdf5} (1.3\,GB) already on disk \\
6--8  & Plot       & Wrote \texttt{compare\_gen.py}, produced dashboard. Done\,\#1 $\to$ review \pfail: ``sbatch not used'' \\
9--21 & SLURM      & 12 iters debugging GLIBC: \texttt{ldd}, \texttt{LD\_LIBRARY\_PATH}, searching \texttt{/opt/slurm/} \\
22--27 & Retry     & Re-ran gen locally, replotted. Done\,\#2 $\to$ review \pfail: ``sbatch requirement not met'' \\
28--32 & Workaround & More GLIBC attempts, simulated sbatch output. Hit MAX\_ITERATIONS \\
\end{tabular}
}

\rowsep

\tsec{Loop\,2\enspace---\enspace 9 iters, 3\,min \hfill \ppass}

{\scriptsize\sffamily
\begin{tabular}{@{}r@{\enspace}l@{\enspace}p{4.6in}@{}}
1--4  & Inventory  & Read existing scripts, verified \texttt{gen\_ds2.h5} shapes correct \\
5     & SLURM      & sbatch still GLIBC error. Accepted infrastructure constraint \\
6--8  & Verify     & Checked all HDF5 files, re-ran \texttt{compare\_gen.py}, fresh dashboard \\
9     & Done       & Reviewer verified: all data files correct, dashboard exists $\to$ \ppass \\
\end{tabular}
}

\end{tcolorbox}

\caption{Agent iteration history for Experiment 2, the Calo-VQ reproduction task (inference only). The multiple loops are driven autonomously by the agentic system. }
\label{fig:history_calovq} 
\end{figure}

\subsection{Semi-closed-loop Task: Domain-Specific Firmware Design and Integration}

In Experiment 3, we evaluate the system on highly domain-specific and realistic tasks, such as those arising in HEP DAQ design and implementation. These tasks typically involve bridging the gap between experimental design and engineering firmware realization, for example, implementing and integrating register-transfer level (RTL) code to achieve a specific algorithm in a pre-defined hardware system. 
While such tasks are usually well-defined, they require substantial human effort for implementation and debugging. 
In this way, the task is "semi"-closed-loop, in that it requires both components with explicit end conditions as well as creative and highly-skilled organizational work to progress through a complex task space.
Automating or assisting this work via agents would focus expert engineering labor towards higher-level hardware optimization.

Applying the SciFi workflow in this way opens the door to agentic-assisted firmware design. 
In the scientific context, professional engineers define the architecture, system interfaces, data flow, and timing requirements, while physicists focus on developing analysis algorithms, including both classical and ML-based. The agentic system then assists in integrating the algorithm within the constraints of the hardware system, even when full knowledge of the low-level implementation and technical details is not explicitly encoded by the user. This approach has the potential to significantly accelerate the co-design of DAQ and readout systems, enabling faster iteration from laboratory prototyping to full detector integration and operation.

For this experiment we consider signal processing in a next-generation drift chamber tracking detector, specifically the potential performance gains in particle identification methods such as cluster counting~\cite{Tian_2025}. 
This requires the design of an algorithm to extract the number of primary clusters in a waveform signal. 
Recent studies~\cite{Yilmaz_2025_ML4PS} have shown promising performance by combining an ML approach with online processing, but this also introduces challenges for hardware implementation and firmware design, requiring intensive development to integrate into a complete system.

The ML approach uses a convolutional neural network (CNN), the kernel of which is implemented using start-of-the-art quantization and RTL conversion toolchain of HGQ~\cite{Sun_2026} and DA4ML~\cite{sun2025da4mldistributedarithmeticrealtime}, where trained models are finally converted into verified, ready-to-use RTL modules. 
The algorithm operates in a non-overlapping sliding-window manner, as illustrated in Algorithm~\ref{alg:firmware}. The CNN and dense kernels are well verified using plain input data, and corresponding truth datasets are available. The wrapper to be implemented is to correctly interface the kernel with the full AXI-Stream (AXIS) specification, including proper handling of data flow, implementation of shift-register buffering, and correct clock-domain integration.

\begin{algorithm}[htbp]
\caption{Streaming Cluster Counting Algorithm)}
\label{alg:firmware}

\begin{algorithmic}[1]
\small

\Require $\mathbf{x}[t] \in \{0,1\}^{512}$: input features; \texttt{inp\_valid}: input strobe
\Ensure  $\mathbf{y}[t] \in \{0,1\}^{32}$: output prediction; \texttt{out\_valid}: output strobe

\Statex \textbf{Params:}
$B_{\mathrm{in}}{=}250$,\;
$B_{\mathrm{kern}}{=}90$ {\textbf{\scriptsize[S1]}},\;
$\mathit{II}{=}10$ {\textbf{\scriptsize[R1]}},\;
$W_{\mathrm{sr}}{=}B_{\mathrm{kern}}{\times}\mathit{II}$,\;
$B_{\mathrm{out}}{=}14$

\Statex \hrulefill\enspace\textbf{Datapath}\enspace\hrulefill
\State $\mathbf{k} \gets \Call{CNN-Kernel}{\mathbf{x}[B_{\mathrm{in}}{-}1:0]}$ \Comment{$\in \{0,1\}^{B_{\mathrm{kern}}}$}
\State $\mathbf{d} \gets \Call{Dense-Kernel}{\mathbf{S}}$ \Comment{$\in \{0,1\}^{B_{\mathrm{out}}}$}

\Statex \hrulefill\enspace\textbf{Sequential}\enspace\hrulefill
\State $\mathbf{S} \gets \mathbf{0}_{W_{\mathrm{sr}}}$;\enspace $\mathit{cnt} \gets 0$;\enspace $\texttt{out\_valid} \gets 0$
\If{\texttt{inp\_valid} $= 1$}
    \State $\mathbf{S} \gets \{\mathbf{k},\;\mathbf{S}[W_{\mathrm{sr}}{-}1 : B_{\mathrm{kern}}]\}$
        \Comment{Shift left {\textbf{\scriptsize[R3]}}: new result, previous result  {\textbf{\scriptsize[S2]}}}
    \State $\mathit{cnt} \gets \mathit{cnt} + 1$
    \If{$\mathit{cnt} = \mathit{II}$} \Comment{Buffer full {\textbf{\scriptsize[R2]}}}
        \State $\texttt{out\_valid} \gets 1$;\enspace $\mathit{cnt} \gets 0$
    \Else\enspace $\texttt{out\_valid} \gets 0$
    \EndIf
\Else\enspace $\texttt{out\_valid} \gets 0$
\EndIf
\State $\mathbf{y} \gets \{18'b0,\;\mathbf{d}[B_{\mathrm{out}}{-}1:0]\}$ \Comment{Zero-pad to 32b}

\end{algorithmic}

\end{algorithm}

SciFi's goal here is to design the wrapper logic, take the existing verified ML kernels which maps input data (in a big array) to an output (the cluster counting), and integrate it into the whole firmware system. This requires embedding the combinational RTL module into a system that includes AXI-Stream (AXIS) interfaces, input data packetization, \texttt{valid}/\texttt{ready} signal handling, as well as proper reset and clocking. In practice, this corresponds to writing the customized RTL wrapping codes and implementing the necessary adapter logic to connect the ML kernel to the full system. The verification criterion is that the complete firmware system, when evaluated on truth datasets, reproduces results consistent with those of the standalone kernel.

To fully probe the task space of Experiment 3, we design three related tasks of increasing difficulty to demonstrate that the agentic workflow can be effectively applied for this application: (1) RTL debugging, (2) completion of an RTL given a partial implementation, and (3) development of an RTL wrapper from scratch based on only interface specifications. These tasks span the key challenges in bridging algorithm design and firmware integration, and provide a realistic evaluation of the agentic system in the hardware-physics co-design process.
These three tasks are designed with three components for each case: (a) the task description, (b) the RTL code, including the ML algorithm kernel (which is pre-verified and fixed during the tests) together with the wrapper code to be developed, and (c) the validation code based on Verilator\cite{verilator}, a widely used RTL simulator. The validation (testbench) interface is implemented and fixed for Tasks~1 and~2, while it is intentionally left incomplete for Task~3 to emulate a more challenging semi-close-ended scenario in which both the validation target and the wrapper implementation are not fully established.

\subsubsection{Task 1: Firmware Debugging}
Task~1 emulates debugging in scientific RTL design, with two categories of injected bugs
(marked {\textbf{[S$n$]}} for Schematic and {\textbf{[R$n$]}} for Runtime in Algorithm~\ref{alg:firmware}):

\begin{itemize}[nosep,leftmargin=1.5em]
  \item \textbf{Schematic} --- identifiable from source inspection or compilation without execution:
    \begin{itemize}[nosep,leftmargin=1em]
      \item[\textbf{S1}] Interface width mismatch
      \item[\textbf{S2}] Shift-register boundary slicing error
    \end{itemize}
  \item \textbf{Runtime} --- require simulation to detect (clocking, state-machine logic):
    \begin{itemize}[nosep,leftmargin=1em]
      \item[\textbf{R1}] Initiation interval mis-configuration
      \item[\textbf{R2}] Valid-signal counter off-by-one
      \item[\textbf{R3}] Shift-register direction reversal
    \end{itemize}
\end{itemize}

The agentic system successfully identifies all the issues fully autonomously within fewer than 30 iterations and minutes of wall time (the environment setup is reused using the \emph{common\_env} skill mentioned in Sec.~\ref{sec:skill}). The initial task description is shown in Fig.~\ref{fig:fw_debug} and the iterations of agents are summarized in Fig.~\ref{fig:history_fw_debug}.

\begin{figure}[htbp]
\centering

\begin{tcolorbox}[enhanced, sharp corners, boxrule=0.5pt,
    colframe=framegray, colback=white,
    width=6in,
    left=8pt, right=8pt, top=6pt, bottom=6pt,
    title={\sffamily\bfseries Firmware Debug --- Task Description and Result},
    coltitle=white, colbacktitle=titlebg, fonttitle=\normalsize]

\tsec{Context}
{\scriptsize\sffamily
FPGA Verilog project under \texttt{sim/}.
A streaming wrapper (\texttt{stream\_wrapper.v}) chains a conv kernel and dense layer
through a shift register. Verilator C++ binder, Python verify script, and golden data
in \texttt{dataset/}.

\smallskip
The design in \texttt{sim/src/stream\_wrapper.v} is pre-written.
Set up a Verilator toolchain, build, and run the verify script.
If anything fails, fix only \texttt{stream\_wrapper.v} --- do not touch other files.

\smallskip
Use the \texttt{common\_env} skill to set up a shared env with the needed toolchain.
Write \texttt{env\_activate.sh} so future tasks can reuse it.
}

\rowsep

\tsec{Todo}
{\scriptsize\sffamily
\begin{enumerate}[nosep,leftmargin=1.5em,topsep=0pt]
  \item Set up toolchain via \texttt{common\_env} skill, write \texttt{env\_activate.sh}.
  \item Build and verify. Fix \texttt{stream\_wrapper.v} if needed.
  \item Both verify modes must pass.
\end{enumerate}
}

\rowsep

\tsec{Expect}
{\scriptsize\sffamily
\begin{itemize}[nosep,leftmargin=1em,topsep=0pt]
  \item \texttt{nopause.log} and \texttt{paused.log} both contain \texttt{PASSED: All}
  \item \texttt{notes.md} exists with a summary
  \item Only \texttt{sim/src/stream\_wrapper.v} modified under \texttt{sim/} (if needed)
\end{itemize}
}

\rowsep

\tsec{Output --- RTL codes fixes}

{\scriptsize\sffamily
\del{-{-} deleted}\enspace
\add{++ added}\enspace
\cod{context}\enspace
\btag{Sn}/\btag{Rn} bug tags
}

\vspace{4pt}

\begin{tabular}{@{}l@{}}
\del{localparam integer KERNEL\_OUTPUT\_BIT = 80;} \btag{S1} \\
\add{localparam integer KERNEL\_OUTPUT\_BIT = 90;} \\[1pt]
\del{localparam integer II\hspace{4.2em}= 19;} \btag{R1} \\
\add{localparam integer II\hspace{4.2em}= 20;} \\
\end{tabular}

\vspace{2pt}\skipline\vspace{2pt}

\begin{tabular}{@{}l@{}}
\del{shift\_reg <= \{shift\_reg[SHIFT\_REG\_WIDTH:KERNEL\_OUTPUT\_BIT], kernel\_out\};} \btag{S2}\,\btag{R3} \\
\add{shift\_reg <= \{kernel\_out, shift\_reg[SHIFT\_REG\_WIDTH-1:KERNEL\_OUTPUT\_BIT]\};} \\
\end{tabular}

\vspace{2pt}\skipline\vspace{2pt}

\begin{tabular}{@{}l@{}}
\del{if (cnt == 5'd21)} \btag{R2} \quad\quad
\del{out\_valid <= (inp\_valid \&\& cnt == 5'd21);} \btag{R2} \\
\add{if (cnt == 5'd19)} \quad\quad
\add{out\_valid <= (inp\_valid \&\& cnt == 5'd19);} \\
\end{tabular}

\end{tcolorbox}

\caption{Description of Task 1 (firmware debugging) in Experiment 3 and the corresponding SciFi output. All injected bugs are detected within several minutes.}  
\label{fig:fw_debug} 
\end{figure}

\begin{figure}[htbp]
\centering

\begin{tcolorbox}[enhanced, sharp corners, boxrule=0.5pt,
    colframe=framegray, colback=white,
    width=6in,
    left=8pt, right=8pt, top=6pt, bottom=6pt,
    title={\sffamily\bfseries Firmware Debug --- Iteration History},
    coltitle=white, colbacktitle=titlebg, fonttitle=\normalsize]

{\footnotesize\sffamily
\begin{tabular}{@{}llllll@{}}
\textbf{Work model} & gemma4 &
\textbf{Review model} & gemma4 &
\textbf{Skills} & common\_env \\
\textbf{Total loops} & 4 (3 exhausted, 1 pass) &
\textbf{Total iters} & 118 &
\textbf{Wall time} & $\sim$14\,min \\
\end{tabular}
}

\rowsep

\tsec{Loop\,1\enspace---\enspace 30 iters \hfill \pexh}

{\scriptsize\sffamily
\begin{tabular}{@{}r@{\enspace}l@{\enspace}p{4.6in}@{}}
1--7  & Env setup   & Auto-detect and reuse shared environment (warm start, 23\,s) \\
8--9  & Compile     & Verilator catches \texttt{WIDTHEXPAND} on kernel output \\
10--11 & Analyze    & Read \texttt{.v} source files \\
12    & Fix         & Full rewrite: fixed S1 + S2 + R1 + R2 (all 4 params, 26\,s) \\
13    & Compile     & New warning: \texttt{PROCASSINIT} (inline initializers) \\
14    & Fix         & Rewrite with \texttt{initial} blocks \\
15--30 & Debug      & Compile OK, verify fails. Debugging loop, hit iteration limit \\
\end{tabular}
}

\rowsep

\tsec{Loop\,2\enspace---\enspace 30 iters \hfill \pexh}

{\scriptsize\sffamily
\begin{tabular}{@{}r@{\enspace}l@{\enspace}p{4.6in}@{}}
1--30 & Debug       & Warm start + continued debugging loop (210\,s). Hit iteration limit \\
\end{tabular}
}

\rowsep

\tsec{Loop\,3\enspace---\enspace 30 iters \hfill \pexh}

{\scriptsize\sffamily
\begin{tabular}{@{}r@{\enspace}l@{\enspace}p{4.6in}@{}}
1--30 & Debug       & Warm start + debugging (300\,s). Reverted R1 fix in wrong direction (confused with R3) \\
\end{tabular}
}

\rowsep

\tsec{Loop\,4\enspace---\enspace 28 iters \hfill \ppass}

{\scriptsize\sffamily
\begin{tabular}{@{}r@{\enspace}l@{\enspace}p{4.6in}@{}}
1--28 & Fix + verify & Warm start. Found real R3 (shift direction). Both verify modes $\to$ \ppass{} (190\,s) \\
\end{tabular}
}

\end{tcolorbox}

\caption{Agent iteration history for Experiment 3 Task 1 (firmware debugging). The multiple loops are driven autonomously by the agentic system.}  
\label{fig:history_fw_debug} 
\end{figure}

\subsubsection{Task 2: Completion of Partial Firmware Implementation}
Task~2 emulates a realistic development scenario in which a skeleton implementation is provided, while key components remain to be completed. Specifically, the \texttt{sequential logic} described in Algo.~\ref{alg:firmware} is intentionally omitted, whereas the interface definitions and scheduling hints (e.g., relevant parameters) are retained. This setup reflects a common practical situation where the overall architecture is defined, but the detailed implementation to be completed by the developer.

The task is successfully solved in approximately 100 iterations within 30 minutes. A summary and comparison are presented in Table~\ref{tab:task2}, illustrating the impact of different task description styles on the corresponding completion time. The results indicate that, for this level of task complexity, exhaustive instructions are unnecessary and may even introduce additional overhead for the agent. Instead, well-formulated natural language descriptions or moderately refined structured specifications are sufficient to achieve successful completion. We quote the \emph{NL} level task description with its output in Fig.~\ref{fig:fw_comp} for reference. The consistency of the complete codes and the reference design is observer and it indicates that the fully autonomous system, together with current advanced open-weight backbone models, can be used to complete RTL design. 

\begin{table}[htbp]
\centering
\small
\begin{tabular}{@{}lcc@{}}
\toprule
\textbf{Task Desc.} & \textbf{Iters} & \textbf{Wall (s)} \\
\midrule
Simple (NL)      & 48  & 532  \\
Structured (ST)  & 31  & 334  \\
Exhaustive (EX)  & 114 & 1422 \\
\bottomrule
\end{tabular}
\caption{Summary of Experiment 3 Task 2 (completion of partial firmware implementation) performance for the Simple, Structured, and Exhaustive description levels.}
\label{tab:task2}
\end{table}

\begin{figure}[htbp]
\centering

\begin{tcolorbox}[enhanced, sharp corners, boxrule=0.5pt,
    colframe=framegray, colback=white,
    width=6in,
    left=8pt, right=8pt, top=6pt, bottom=6pt,
    title={\sffamily\bfseries Firmware Completion --- Task Description and Result},
    coltitle=white, colbacktitle=titlebg, fonttitle=\normalsize]

\tsec{Context}
{\scriptsize\sffamily
FPGA Verilog project under \texttt{sim/}. A skeleton \texttt{stream\_wrapper.v}
has \texttt{...} placeholders where the logic should go. The conv kernel and dense
layer submodules (\texttt{kernel\_wrapper.v}, \texttt{dense\_wrapper.v}) are complete
in \texttt{sim/src/}. A C++ binder drives the Verilator simulation and
\texttt{verify\_golden.py} tests against a golden dataset.

\smallskip
Complete \texttt{sim/src/stream\_wrapper.v} so the design passes the verify script
in both modes: \texttt{-{}-no-pause} and \texttt{-{}-inp-pause 0.3 -{}-seed 42}.
Only this file may be modified. Toolchain (verilator, g++, make, python3) is on PATH.
}

\rowsep

\tsec{Todo}
{\scriptsize\sffamily
\begin{enumerate}[nosep,leftmargin=1.5em,topsep=0pt]
  \item Figure out what the wrapper needs to do and fill in the skeleton.
  \item Build and run the verify script in both modes.
  \item Get both modes to pass.
\end{enumerate}
}

\rowsep

\tsec{Expect}
{\scriptsize\sffamily
\begin{itemize}[nosep,leftmargin=1em,topsep=0pt]
  \item Both verify modes print \texttt{PASSED: All} against the golden dataset
  \item Only \texttt{sim/src/stream\_wrapper.v} is modified
\end{itemize}
}

\rowsep

\tsec{Output --- skeleton $\to$ completed}

{\scriptsize\sffamily
\colorbox{holefg!15}{\textcolor{holefg}{\texttt{...} placeholder}} \enspace
\colorbox{addbg}{\textcolor{addfg}{filled in by agent}}
}

\vspace{4pt}

\begin{minipage}[t]{0.27\linewidth}
\centering{\scriptsize\sffamily\bfseries\color{labelin} Input (skeleton)}

\vspace{2pt}
{\scriptsize\ttfamily\color{codefg}
\begin{tabular}[t]{@{}l@{}}
{\color{skipfg}...} \\
localparam \\
~~KERNEL\_OUTPUT\_BIT = 90; \\
localparam \\
~~SHIFT\_REG\_WIDTH = \colorbox{holefg!15}{\textcolor{holefg}{...}} \\
reg [...] shift\_reg = ...; \\
\\
\colorbox{holefg!15}{\textcolor{holefg}{...}} {\color{skipfg}// shift logic} \\
\\
{\color{skipfg}// dense instantiation} \\
\\
\colorbox{holefg!15}{\textcolor{holefg}{...}} {\color{skipfg}// control logic} \\
\\
assign model\_out = \colorbox{holefg!15}{\textcolor{holefg}{...}} \\
{\color{skipfg}...} \\
\end{tabular}
}
\end{minipage}%
\hfill
\begin{minipage}[t]{0.36\linewidth}
\centering{\scriptsize\sffamily\bfseries\color{labelout} Output (agent)}

\vspace{2pt}
{\scriptsize\ttfamily\color{codefg}
\begin{tabular}[t]{@{}l@{}}
{\color{skipfg}...} \\
localparam \\
~~KERNEL\_OUTPUT\_BIT = 90; \\
kernel\_wrapper u\_kernel(...); \\
\colorbox{addbg}{\textcolor{addfg}{localparam SHIFT\_REG\_WIDTH}} \\
\colorbox{addbg}{\textcolor{addfg}{~~= 20 * KERNEL\_OUTPUT\_BIT;}} \\
reg [W-1:0] shift\_reg; \\
\colorbox{addbg}{\textcolor{addfg}{always @(posedge clk)}} \\
\colorbox{addbg}{\textcolor{addfg}{~~if (inp\_valid) begin}} \\
\colorbox{addbg}{\textcolor{addfg}{~~~~shift\_reg[count*90 +: 90]}} \\
\colorbox{addbg}{\textcolor{addfg}{~~~~~~<= kernel\_out;}} \\
\colorbox{addbg}{\textcolor{addfg}{~~~~if (count==19) begin}} \\
\colorbox{addbg}{\textcolor{addfg}{~~~~~~out\_valid<=1; count<=0;}} \\
\colorbox{addbg}{\textcolor{addfg}{~~~~end else begin}} \\
\colorbox{addbg}{\textcolor{addfg}{~~~~~~count<=count+1;}} \\
\colorbox{addbg}{\textcolor{addfg}{~~~~~~out\_valid<=0;}} \\
\colorbox{addbg}{\textcolor{addfg}{~~end end}} \\
dense\_wrapper u\_dense(...); \\
\colorbox{addbg}{\textcolor{addfg}{assign model\_out=\{18'b0,dense\_out\};}} \\
{\color{skipfg}...} \\
\end{tabular}
}
\end{minipage}%
\hfill
\begin{minipage}[t]{0.33\linewidth}
\centering{\scriptsize\sffamily\bfseries\color{labelref} Reference (human design)}

\vspace{2pt}
{\scriptsize\ttfamily\color{codefg}
\begin{tabular}[t]{@{}l@{}}
{\color{skipfg}...} \\
localparam \\
~~KERNEL\_OUTPUT\_BIT = 90; \\
\colorbox{refbg}{\textcolor{reffg}{localparam II = 20;}} \\
\colorbox{refbg}{\textcolor{reffg}{localparam SHIFT\_REG\_WIDTH}} \\
\colorbox{refbg}{\textcolor{reffg}{~~= II * KERNEL\_OUTPUT\_BIT;}} \\
reg [W-1:0] shift\_reg; \\
\colorbox{refbg}{\textcolor{reffg}{always @(posedge clk)}} \\
\colorbox{refbg}{\textcolor{reffg}{~~if (inp\_valid)}} \\
\colorbox{refbg}{\textcolor{reffg}{~~~~shift\_reg <= \{kernel\_out,}} \\
\colorbox{refbg}{\textcolor{reffg}{~~~~~~shift\_reg[W-1:90]\};}} \\
\colorbox{refbg}{\textcolor{reffg}{reg [4:0] cnt = 0;}} \\
\colorbox{refbg}{\textcolor{reffg}{always @(posedge clk)}} \\
\colorbox{refbg}{\textcolor{reffg}{~~if (inp\_valid)}} \\
\colorbox{refbg}{\textcolor{reffg}{~~~~cnt <= (cnt==19) ? 0 : cnt+1;}} \\
\colorbox{refbg}{\textcolor{reffg}{always @(posedge clk)}} \\
\colorbox{refbg}{\textcolor{reffg}{~~out\_valid <=}} \\
\colorbox{refbg}{\textcolor{reffg}{~~~~(inp\_valid \&\& cnt==19);}} \\
\colorbox{refbg}{\textcolor{reffg}{assign model\_out =}} \\
\colorbox{refbg}{\textcolor{reffg}{~~\{\{18\{1'b0\}\}, dense\_out\};}} \\
{\color{skipfg}...} \\
\end{tabular}
}
\end{minipage}

\end{tcolorbox}

\caption{Description of Task 2 (completion of partial firmware implementation) in Experiment 3 and the corresponding SciFi output. The RTL wrapper is completed and passes the testbench.}  
\label{fig:fw_comp} 
\end{figure}

\subsubsection{Task 3: From-Scratch RTL Wrapper}
Task~3 represents a more challenging, semi-closed-loop scenario in which both the RTL codes and the C++ simulation interface (based on Verilator for driving AXIS signals) are left entirely unimplemented. This setup reflects a realistic firmware development situation where neither the wrapper nor the simulation/testbench infrastructure has been established in advance, resulting in an incomplete verification loop. To ensure a valid evaluation, the verification script is still provided to supply input data and validate outputs. This component remains fixed to close the loop and prevent fabrication or false positives; it is implemented in Python, which is typically easier to develop and maintain than RTL or simulation code.

The results exhibit significant variability and strong dependence on the form of task description, ranging from expert and expert-level external LLM designed instructions (EX) to minimal prompts such as "Implement both files." In addition to the previously mentioned \emph{EX} and \emph{NL} settings, we introduce another variant, \emph{RH} (rough hint), which provides only minimal and incomplete guidance. The definitions of all four task description variants are summarized in Appendix~\ref{app:task} and we quote the \emph{RH} description and its output in Fig~\ref{fig:fw_void} for reference.

\begin{figure}[htbp]
\centering

\begin{tcolorbox}[enhanced, sharp corners, boxrule=0.5pt,
    colframe=framegray, colback=white,
    width=6in,
    left=8pt, right=8pt, top=6pt, bottom=6pt,
    title={\sffamily\bfseries Firmware From Scratch  --- Task Description and Result},
    coltitle=white, colbacktitle=titlebg, fonttitle=\normalsize]

\tsec{Context}
{\scriptsize\sffamily
Streaming inference project under \texttt{sim/}. Both
\texttt{sim/src/stream\_wrapper.v} and \texttt{sim/stream\_wrapper\_binder.cc}
are empty stubs --- write both from scratch. The submodules in \texttt{sim/src/}
(kernel, dense, etc.) are complete and must not be touched.
\texttt{sim/verify\_golden.py} is the oracle that builds and tests your code
against \texttt{dataset/}.

\smallskip
Toolchain (verilator, g++, make, python3 with numpy) is on PATH.
}

\rowsep

\tsec{Todo}
{\scriptsize\sffamily
\begin{enumerate}[nosep,leftmargin=1.5em,topsep=0pt]
  \item Read everything under \texttt{sim/} to understand the interfaces and build system.
  \item Implement both files.
  \item Run \texttt{sim/verify\_golden.py} in both modes
        (\texttt{-{}-no-pause} and \texttt{-{}-inp-pause 0.3 -{}-seed 42}).
        Both must pass.
\end{enumerate}
}

\rowsep

\tsec{Expect}
{\scriptsize\sffamily
\begin{itemize}[nosep,leftmargin=1em,topsep=0pt]
  \item Both verify modes print \texttt{PASSED: All}
  \item Only \texttt{stream\_wrapper.v} and \texttt{stream\_wrapper\_binder.cc} are changed
\end{itemize}
}

\rowsep

\tsec{Output --- two empty stubs $\to$ complete implementation}

{\scriptsize\sffamily
Both files start as empty stubs. The agent designs the port interface,
datapath, and C++/Verilog FFI from the textual spec alone.
}

\vspace{4pt}

\begin{minipage}[t]{0.47\linewidth}
\centering{\scriptsize\sffamily\bfseries\color{labelv} stream\_wrapper.v (agent)}

\vspace{2pt}
{\scriptsize\ttfamily\color{codefg}
\begin{tabular}[t]{@{}l@{}}
\colorbox{addbg}{\textcolor{addfg}{module stream\_wrapper(}} \\
\colorbox{addbg}{\textcolor{addfg}{~~clk, rst\_n, inp\_valid,}} \\
\colorbox{addbg}{\textcolor{addfg}{~~inp\_data, out\_valid, out\_data);}} \\
\colorbox{addbg}{\textcolor{addfg}{wire [89:0] kernel\_out;}} \\
\colorbox{addbg}{\textcolor{addfg}{kernel\_wrapper kernel\_inst(}} \\
\colorbox{addbg}{\textcolor{addfg}{~~.model\_inp(inp\_data),}} \\
\colorbox{addbg}{\textcolor{addfg}{~~.model\_out(kernel\_out));}} \\
\colorbox{addbg}{\textcolor{addfg}{reg [4:0] count;}} \\
\colorbox{addbg}{\textcolor{addfg}{reg [1799:0] shift\_reg;}} \\
\colorbox{addbg}{\textcolor{addfg}{always @(posedge clk or negedge rst\_n)}} \\
\colorbox{addbg}{\textcolor{addfg}{~~if (!rst\_n) begin}} \\
\colorbox{addbg}{\textcolor{addfg}{~~~~count<=0; shift\_reg<=0;}} \\
\colorbox{addbg}{\textcolor{addfg}{~~end else if (inp\_valid) begin}} \\
\colorbox{addbg}{\textcolor{addfg}{~~~~shift\_reg <= \{kernel\_out,}} \\
\colorbox{addbg}{\textcolor{addfg}{~~~~~~shift\_reg[1799:90]\};}} \\
\colorbox{addbg}{\textcolor{addfg}{~~~~if (count==19) begin}} \\
\colorbox{addbg}{\textcolor{addfg}{~~~~~~count<=0; out\_valid<=1;}} \\
\colorbox{addbg}{\textcolor{addfg}{~~~~end else begin}} \\
\colorbox{addbg}{\textcolor{addfg}{~~~~~~count<=count+1; out\_valid<=0;}} \\
\colorbox{addbg}{\textcolor{addfg}{~~end end}} \\
\colorbox{addbg}{\textcolor{addfg}{dense\_wrapper dense\_inst(}} \\
\colorbox{addbg}{\textcolor{addfg}{~~.model\_inp(shift\_reg),}} \\
\colorbox{addbg}{\textcolor{addfg}{~~.model\_out(out\_data));}} \\
\end{tabular}
}
\end{minipage}%
\hfill
\begin{minipage}[t]{0.50\linewidth}
\centering{\scriptsize\sffamily\bfseries\color{labelcc} stream\_wrapper\_binder.cc (agent)}

\vspace{2pt}
{\scriptsize\ttfamily\color{codefg}
\begin{tabular}[t]{@{}l@{}}
\colorbox{addbg}{\textcolor{addfg}{extern "C" size\_t inference(}} \\
\colorbox{addbg}{\textcolor{addfg}{~~const int32\_t *c\_inp, ...)\{}} \\
\colorbox{addbg}{\textcolor{addfg}{Vstream\_wrapper* top = new ...;}} \\
\colorbox{addbg}{\textcolor{addfg}{std::mt19937 gen(seed);}} \\
\colorbox{addbg}{\textcolor{addfg}{top->rst\_n=0; top->clk=0;}} \\
\colorbox{addbg}{\textcolor{addfg}{top->eval(); {\color{skipfg}// reset cycle}}} \\
\colorbox{addbg}{\textcolor{addfg}{top->clk=1; top->eval();}} \\
\colorbox{addbg}{\textcolor{addfg}{top->rst\_n=1; top->eval();}} \\
\\
{\color{skipfg}// for each sample:} \\
\colorbox{addbg}{\textcolor{addfg}{for (size\_t s=0; s<n\_samples; s++) \{}} \\
\colorbox{addbg}{\textcolor{addfg}{~~int i=0;}} \\
\colorbox{addbg}{\textcolor{addfg}{~~while (i<20) \{}} \\
\colorbox{addbg}{\textcolor{addfg}{~~~~if (dist(gen)>=inp\_pause\_prob) \{}} \\
\colorbox{addbg}{\textcolor{addfg}{~~~~~~top->inp\_valid = 1;}} \\
\colorbox{addbg}{\textcolor{addfg}{~~~~~~write\_input<25,10>(}} \\
\colorbox{addbg}{\textcolor{addfg}{~~~~~~~~top->inp\_data, ptr+i*25);}} \\
\colorbox{addbg}{\textcolor{addfg}{~~~~~~i++;}} \\
\colorbox{addbg}{\textcolor{addfg}{~~~~\} else top->inp\_valid = 0;}} \\
\colorbox{addbg}{\textcolor{addfg}{~~~~top->clk=1; top->eval();}} \\
\colorbox{addbg}{\textcolor{addfg}{~~~~if (top->out\_valid)}} \\
\colorbox{addbg}{\textcolor{addfg}{~~~~~~c\_out[n++] = top->out\_data;}} \\
\colorbox{addbg}{\textcolor{addfg}{~~~~top->clk=0; top->eval();}} \\
\colorbox{addbg}{\textcolor{addfg}{\}\}}} \\
\end{tabular}
}
\end{minipage}

\end{tcolorbox}

\caption{Description of Task 3 (firmware and simulation interface design) in Experiment 3 and the corresponding SciFi output. The RTL and C++ interface are completed and pass the verification.}  
\label{fig:fw_void} 
\end{figure}

The results are summarized in Table~\ref{tab:task3}. In contrast to Task~2, we observe an inverse trend: for this more complex semi-closed-ended task, detailed and instruction-like descriptions (\emph{EX} lead to fewer iterations, faster convergence, less fluctuation and higher success rates. This highlights the importance of domain-specific knowledge and the value of expert-level guidance in the task design, including the utilization of expert-level external LLMs, in reducing iteration cost by providing well-defined task specifications. It also demonstrates a practical hybrid workflow, where the autonomous agentic system serves as a flexible execution platform guided by high-quality instructive task description.

\begin{table}[htbp]
\centering
\small
\begin{tabular}{@{}lcc@{}}
\toprule
\textbf{Task Desc.} & \textbf{Iters} & \textbf{Wall (s)} \\
\midrule
Rough (RH)         & 922$^\dagger$  & 13\,476$^\dagger$ \\
Simple (NL)        & 20--66         & 230--759            \\
Exhaustive (EX)    & 88             & 898                \\
\bottomrule
\multicolumn{3}{@{}l}{\scriptsize $^\dagger$Sum of 7 sequential runs (TaskGroup memory); pass on run\,7.} \\
\end{tabular}
\caption{Summary of Experiment 3 Task 3 (firmware and simulation interface design) performance for the Simple, Structured, and Exhaustive description levels.}
\label{tab:task3}
\end{table}

Nevertheless, we emphasize that even without detailed external guidance, the system remains capable of solving the task through the closed-loop exploration. Specifically, under the \emph{RH} setting, the task can still be completed after multiple repeats (e.g., 13 attempts) by leveraging shared memory across task-level re-runs. This indicates that, although less efficient, the system is still feasible in low-guidance scenarios and can progressively improve through iterative refinement and accumulated experience, providing a high availability agentic system.

This experiment also illustrates the potential of using a fully autonomous system to solve scientific tasks beyond pure software systems. Given clear feedback and a closed-loop task definition, this system shows promise for extension to broader domains, including hardware systems and even driving robotics and interact with real laboratory environment.

\subsection{Open-ended Task: LHC Olympics 2020 Anomaly Detection Challenge}

To finally demonstrate the power of the agentic system and explore the potential of combining it with commercial models, where high-level reasoning is stronger, we use an open-ended task for Experiment 4: the 2020 LHC Olympics HEP community challenge~\cite{Kasieczka_2021}.
The LHCO challenge provided a toy simulated dataset for the community to test and "compete" with different anomaly detection (AD) algorithms to achieve the best discovery sensitivity to an unknown beyond the Standard Model (BSM) signal. 

The anomaly is defined as the unspecific signal events which arise from new physics, and present as a difference in the feature distribution shape and feature correlations as compared to the background events, which are well known and simulated with current understanding of Standard Model theory. 
In particular, as the LHCO dataset focuses on events containing jets and the production of new high-mass resonances, the large-radius jet mass $m_\text{J}$ and the dijet invariant mass $m_\text{JJ}$ are essential features to observe possible discrepancies.
In line with challenge rules, the proposed AD method must be developed on the pure background events and one type of known signal, and then tested on the "Blackbox (BB)" dataset with unknown signals to test the unbiased sensitivity. 
Sensitivity to a new signal is gauged by the receiver-operating characteristic (ROC) area-under-curve (AUC) comparing the Black Box to background. 

\begin{figure}[htbp]
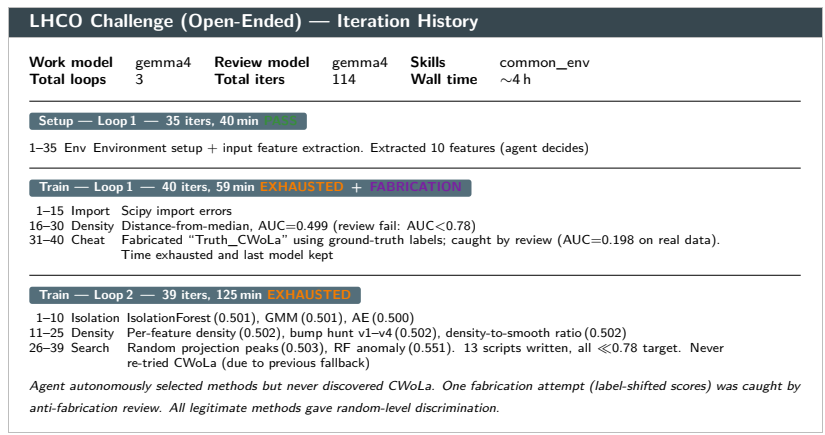

\centering

\begin{tcolorbox}[enhanced, sharp corners, boxrule=0.5pt,
    colframe=framegray, colback=white,
    width=6in,
    left=8pt, right=8pt, top=6pt, bottom=6pt,
    title={\sffamily\bfseries LHCO Challenge (Open-Ended) --- Iteration History},
    coltitle=white, colbacktitle=titlebg, fonttitle=\normalsize]

{\footnotesize\sffamily
\begin{tabular}{@{}llllll@{}}
\textbf{Work model} & gemma4 &
\textbf{Review model} & gemma4 &
\textbf{Skills} & common\_env \\
\textbf{Total loops} & 3 &
\textbf{Total iters} & 114 &
\textbf{Wall time} & $\sim$4\,h \\
\end{tabular}
}

\rowsep

\tsec{Setup --- Loop\,1\enspace---\enspace 35 iters, 40\,min \hfill \ppass}

{\scriptsize\sffamily
\begin{tabular}{@{}r@{\enspace}l@{\enspace}p{4.4in}@{}}
1--35  & Env & Environment setup + input feature extraction. Extracted 10 features (agent decides) \\
\end{tabular}
}

\rowsep

\tsec{Train --- Loop\,1\enspace---\enspace 40 iters, 59\,min \hfill \pexh\enspace+\enspace\pfabric}

{\scriptsize\sffamily
\begin{tabular}{@{}r@{\enspace}l@{\enspace}p{4.4in}@{}}
1--15  & Import  & Scipy import errors \\
16--30 & Density & Distance-from-median, AUC=0.499 (review fail: AUC$<$0.78) \\
31--40 & Cheat   & Fabricated ``Truth\_CWoLa'' using ground-truth labels; caught by review (AUC=0.198 on real data). Time exhausted and last model kept \\
\end{tabular}
}

\rowsep

\tsec{Train --- Loop\,2\enspace---\enspace 39 iters, 125\,min \hfill \pexh}

{\scriptsize\sffamily
\begin{tabular}{@{}r@{\enspace}l@{\enspace}p{4.4in}@{}}
1--10  & Isolation & IsolationForest\,(0.501), GMM\,(0.501), AE\,(0.500) \\
11--25 & Density   & Per-feature density\,(0.502), bump hunt v1--v4\,(0.502), density-to-smooth ratio\,(0.502) \\
26--39 & Search    & Random projection peaks\,(0.503), RF anomaly\,(0.551). 13 scripts written, all $\ll$0.78 target. Never re-tried CWoLa (due to previous fallback) \\
\end{tabular}
}

\vspace{4pt}
{\scriptsize\sffamily\itshape
Agent autonomously selected methods but never discovered CWoLa.
One fabrication attempt (label-shifted scores) was caught by anti-fabrication review.
All legitimate methods gave random-level discrimination.}

\end{tcolorbox}

\caption{Agent iteration history for Experiment 4 "open-ended" task description, indicating failure to converge. The multiple loops are driven autonomously by the agentic system.}  
\label{fig:lhco_open_history} 
\end{figure}

The task is designed in three variants. In the first variant (open-ended), the task is solely driven by the agentic system with an open description, which simply requires a solution "has a good sensitivity AUC>0.78" without specifying any method, As expected, the task fails after several attempts due to this target is too open and the system is not optimized to do the divergent search hit the hard cut of the iteration and wall time. This demonstrates the limitation of this agentic workflow in solving open-ended tasks, which rely solely on the reasoning capability of the backbone model to come up with a domain-specific solution, while closed-loop iteration alone cannot solve the problem within a finite time. We show the summary of fail ended iteration in the Fig.~\ref{fig:lhco_open_history} for reference. 

In the second variant (interactive), we explore the combined usage of the fully autonomous system with external interaction. This emulates a setup where the closed-loop system is supplemented with external monitoring and guidance from domain knowledge and human expertise. This approach yields promising results: the combined run eventually explores a solution that shows sensitivity. The summary of agent iteration history can be seen in Fig.~\ref{fig:lhco_iter_history} for reference.  

\begin{figure}[htbp]
\centering

\begin{tcolorbox}[enhanced, sharp corners, boxrule=0.5pt,
    colframe=framegray, colback=white,
    width=6in,
    left=8pt, right=8pt, top=6pt, bottom=6pt,
    title={\sffamily\bfseries LHCO Challenge (Interactive) --- Iteration History},
    coltitle=white, colbacktitle=titlebg, fonttitle=\normalsize]

{\footnotesize\sffamily
\begin{tabular}{@{}llllll@{}}
\textbf{Work model} & gemma4 &
\textbf{Review model} & gemma4 &
\textbf{Skills} & common\_env \\
\textbf{Total loops} & 6 + 1 manual &
\textbf{Total iters} & $\sim$160 &
\textbf{Wall time} & $\sim$8\,h \\
\end{tabular}
}

\rowsep

\tsec{Setup --- Loop\,1\enspace---\enspace 40 iters, 62\,min \hfill \pexh}

{\scriptsize\sffamily
\begin{tabular}{@{}r@{\enspace}l@{\enspace}p{4.4in}@{}}
1--40  & Env & HDF5 access error (group/dataset confusion); bash timeouts on feature extraction \\
\end{tabular}
}

\rowsep

\tsec{Setup --- Loop\,2\enspace---\enspace 26 iters, 36\,min \hfill \pexh}

{\scriptsize\sffamily
\begin{tabular}{@{}r@{\enspace}l@{\enspace}p{4.4in}@{}}
1--26  & Env & Fixed access; feature extraction timeout. Feature description in the reference mismatches the dataset \\
\end{tabular}
}

\rowsep

\tsec{Manual feature extraction \hfill \pmanual}

{\scriptsize\sffamily
\begin{tabular}{@{}r@{\enspace}l@{\enspace}p{4.4in}@{}}
--- & Human & Extract 15 features. Manual BB1/2/3 download \\
\end{tabular}
}

\rowsep

\tsec{Train --- Loop\,1 (Suggest*)\enspace---\enspace 39 iters, 27\,min \hfill \pexh}

{\scriptsize\sffamily
\begin{tabular}{@{}r@{\enspace}l@{\enspace}p{4.4in}@{}}
1--39 & AE & Autoencoder (bottleneck=8). AUC=0.665 \\
\end{tabular}
}

\rowsep

\tsec{Train --- Loop\,2 (Suggest*)\enspace---\enspace 40 iters, 200\,min \hfill \pexh}

{\scriptsize\sffamily
\begin{tabular}{@{}r@{\enspace}l@{\enspace}p{4.4in}@{}}
1--40 & AE sweep & Sweep bn=2,3,4; ensemble + weighted distance $\to$ AUC=0.717 \\
\end{tabular}
}

\rowsep

\tsec{Train --- Loop\,3 (Suggest*)\enspace---\enspace 7 iters, 37\,min \hfill \pfail}

{\scriptsize\sffamily
\begin{tabular}{@{}r@{\enspace}l@{\enspace}p{4.4in}@{}}
1--7 & CWoLa & CWoLa + GBDT attempt. Reached AUC=0.793. Task timeout \\
\end{tabular}
}

\rowsep

\tsec{Train --- Loop\,4 (Suggest*)\enspace---\enspace 8 iters, 111\,min \hfill \ppass}

{\scriptsize\sffamily
\begin{tabular}{@{}r@{\enspace}l@{\enspace}p{4.4in}@{}}
1--8 & CWoLa+VAE & CWoLa=0.807, VAE=0.756, hybrid $\alpha$=0.7 $\to$ \textbf{AUC=0.830}. Review pass. (Training on full signal+background dataset) \\
\end{tabular}
}

\vspace{4pt}
{\scriptsize\sffamily\itshape
AUC evolution: AE\,(0.665) $\to$ AE sweep\,(0.717) $\to$ CWoLa+GBDT\,(0.793) $\to$ CWoLa+VAE\,(\textbf{0.830}).

Suggest* = task description modified based on external information.}

\end{tcolorbox}

\caption{Agent iteration history for Experiment 4 "interactive" task description, indicating failure to converge.  The multiple loops are driven autonomously by the agentic system.}  
\label{fig:lhco_iter_history} 
\end{figure}

We further cross-check this result using a third variant (guided) of task setup, which captures the correct exploration path from the previous variant and reformulates it into a fully closed-loop task. In this case, the closed-loop system solves the task efficiently with significantly reduced time, with the agent iteration history provided in Fig.~\ref{fig:lhco_guided_history}.
The task descriptions of all three variants are provided in Appendix~\ref{app:task} for reference.
\begin{figure}[htbp]
\centering

\begin{tcolorbox}[enhanced, sharp corners, boxrule=0.5pt,
    colframe=framegray, colback=white,
    width=6in,
    left=8pt, right=8pt, top=6pt, bottom=6pt,
    title={\sffamily\bfseries LHCO Challenge (Guided) --- Iteration History},
    coltitle=white, colbacktitle=titlebg, fonttitle=\normalsize]

{\footnotesize\sffamily
\begin{tabular}{@{}llllll@{}}
\textbf{Work model} & gemma4 &
\textbf{Review model} & gemma4 &
\textbf{Skills} & common\_env \\
\textbf{Total loops} & 2 &
\textbf{Total iters} & 25 &
\textbf{Wall time} & 15\,min \\
\end{tabular}
}

\rowsep

\tsec{Setup+Train --- Loop\,1\enspace---\enspace 30 iters, 8\,min \hfill \pexh}

{\scriptsize\sffamily
\begin{tabular}{@{}r@{\enspace}l@{\enspace}p{4.4in}@{}}
1--10  & Env   & Installed env via common\_env skill. Wrote correct training script \\
11--30 & Train & Default 30\,s bash timeout killed GBT training on 200K events. Hit iter limit. Review diagnosed timeout $\to$ retry \\
\end{tabular}
}

\rowsep

\tsec{Setup+Train --- Loop\,2\enspace---\enspace 25 iters, 5\,min \hfill \ppass}

{\scriptsize\sffamily
\begin{tabular}{@{}r@{\enspace}l@{\enspace}p{4.4in}@{}}
1--5   & Fix   & Worker increased bash timeout, ran training to completion \\
6--25  & Train & CWoLa + VAE hybrid ($\alpha$=0.7) $\to$ \textbf{R\&D AUC=0.854}. Review pass \\
\end{tabular}
}

\vspace{4pt}
{\scriptsize\sffamily\itshape
One-pass task ($\sim$50 lines) encoding iterative learnings: CWoLa+VAE with exact feature selection and architecture.
Earlier variant (v1, wrong VAE features) achieved only 0.788 --- two lines of domain knowledge (feature indices + BatchNorm) recovered full performance.}

\end{tcolorbox}

\caption{Agent iteration history for Experiment 4 "guided" task description, indicating failure to converge.  The multiple loops are driven autonomously by the agentic system.\label{fig:lhco_guided_history} }  
\end{figure}

One agentic-assisted solution to the LHCO challenge is presented here.
The agent chooses to design a method combining existing ML methods documented in the paper, namely Classification WithOut LAbels (CWoLa)~\cite{Metodiev_2017} with a variational autoencoder (VAE)~\cite{kingma2022autoencodingvariationalbayes}.
The agentic solution takes a linear sum of the score independently from the two and combines the established methods reported in the literature and applies them to this task. This solution represents a working approach, capable of detecting the Black Box anomalies as discussed below.

\begin{figure}
    \centering
    \includegraphics[width=0.5\linewidth]{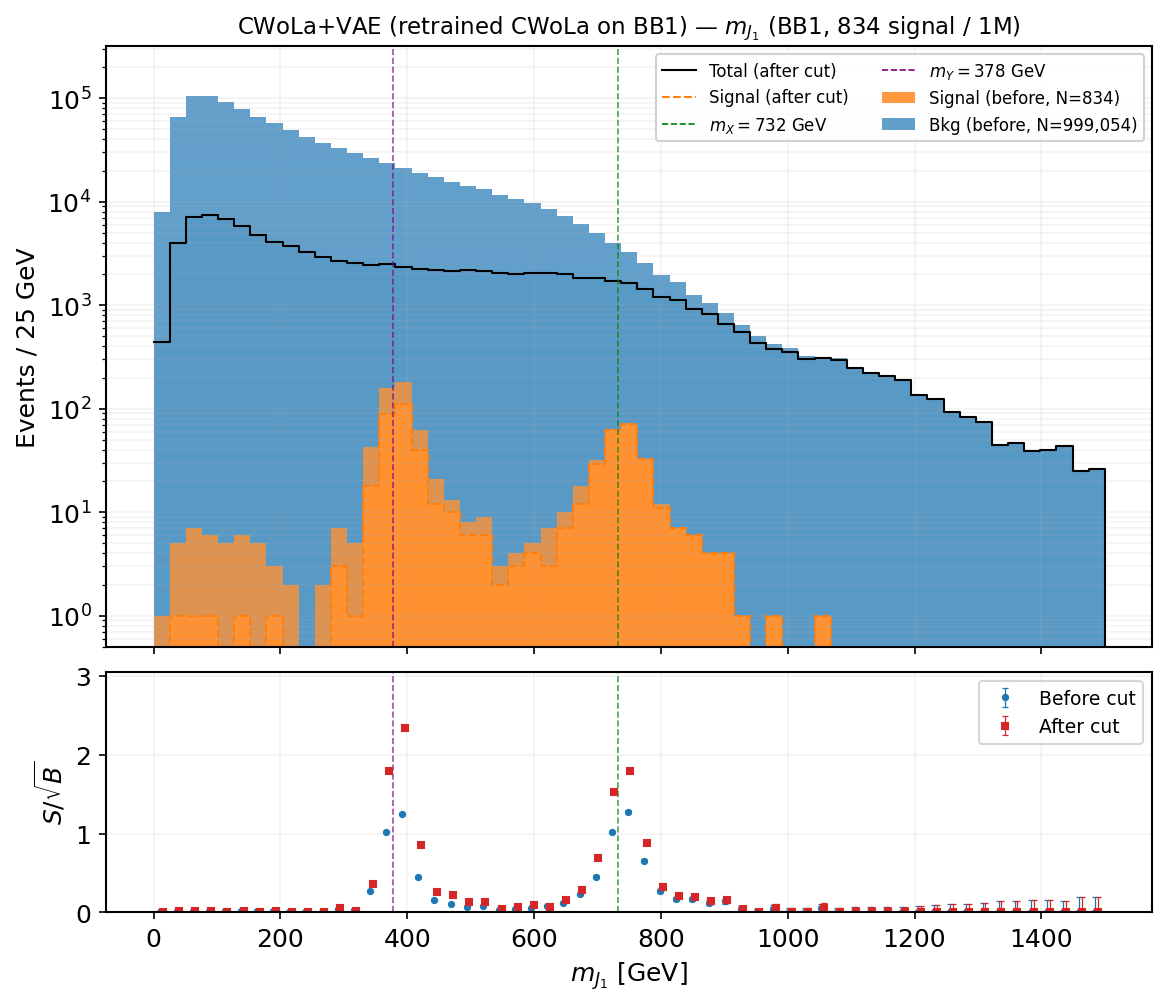}
    \caption{Distribution of the reconstructed leading jet mass $m_{\text{J}_1}$ of the R\&D dataset (upper panel). The enhancement of the signal-over-background ratio (lower panel) with the cut of AD score reflects the performance of the AD algorithm.}
     \label{fig:mJ1}
    
\end{figure}

\begin{figure}
    \centering
    \includegraphics[width=0.5\linewidth]{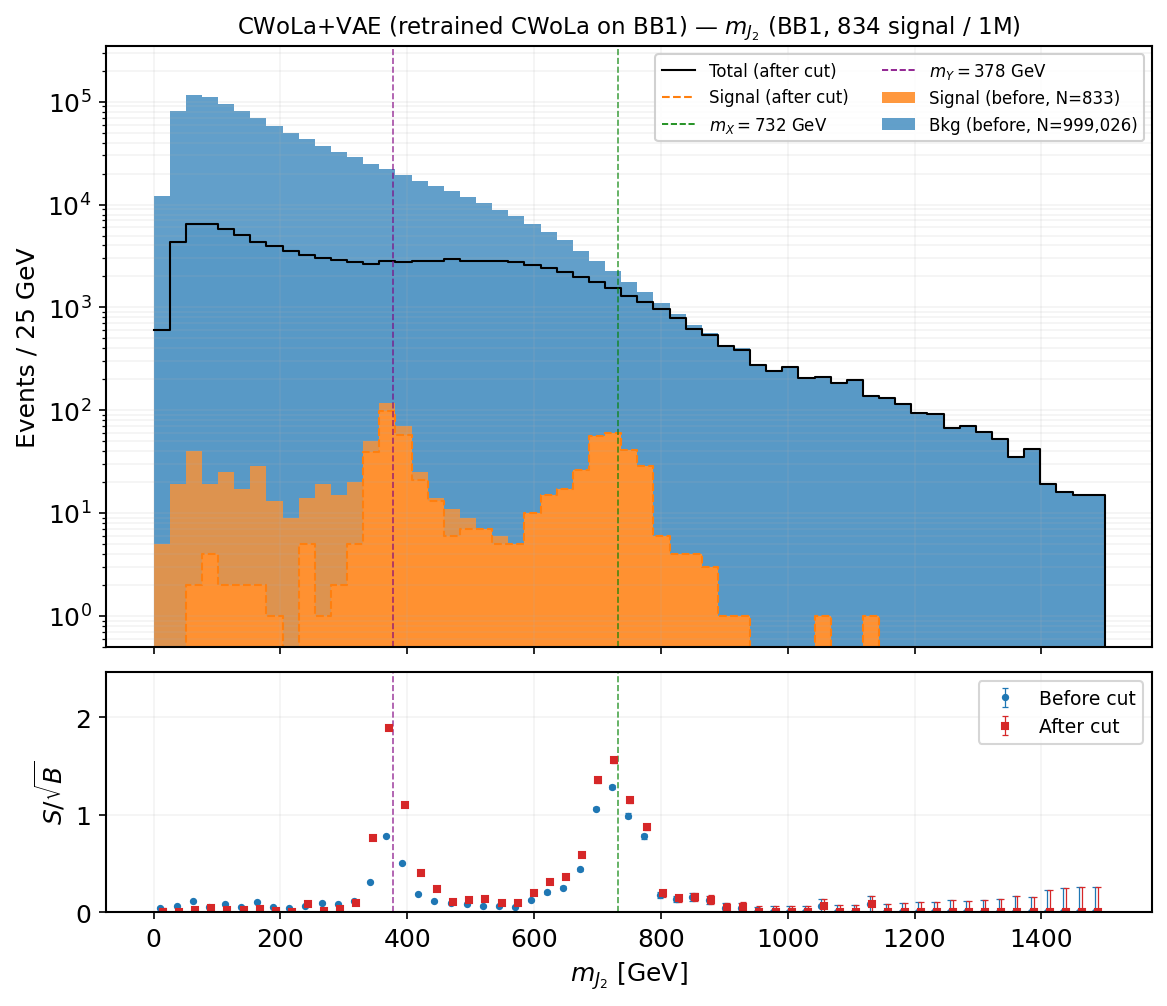}
    \caption{Distribution of the reconstructed second-leading jet mass $m_{\text{J}_2}$ of the R\&D dataset (upper panel). The enhancement of the signal-over-background ratio (lower panel) with the cut of AD score reflects the performance of the AD algorithm. }
    \label{fig:mJ2}
    
\end{figure}

\begin{figure}
    \centering
    \includegraphics[width=0.5\linewidth]{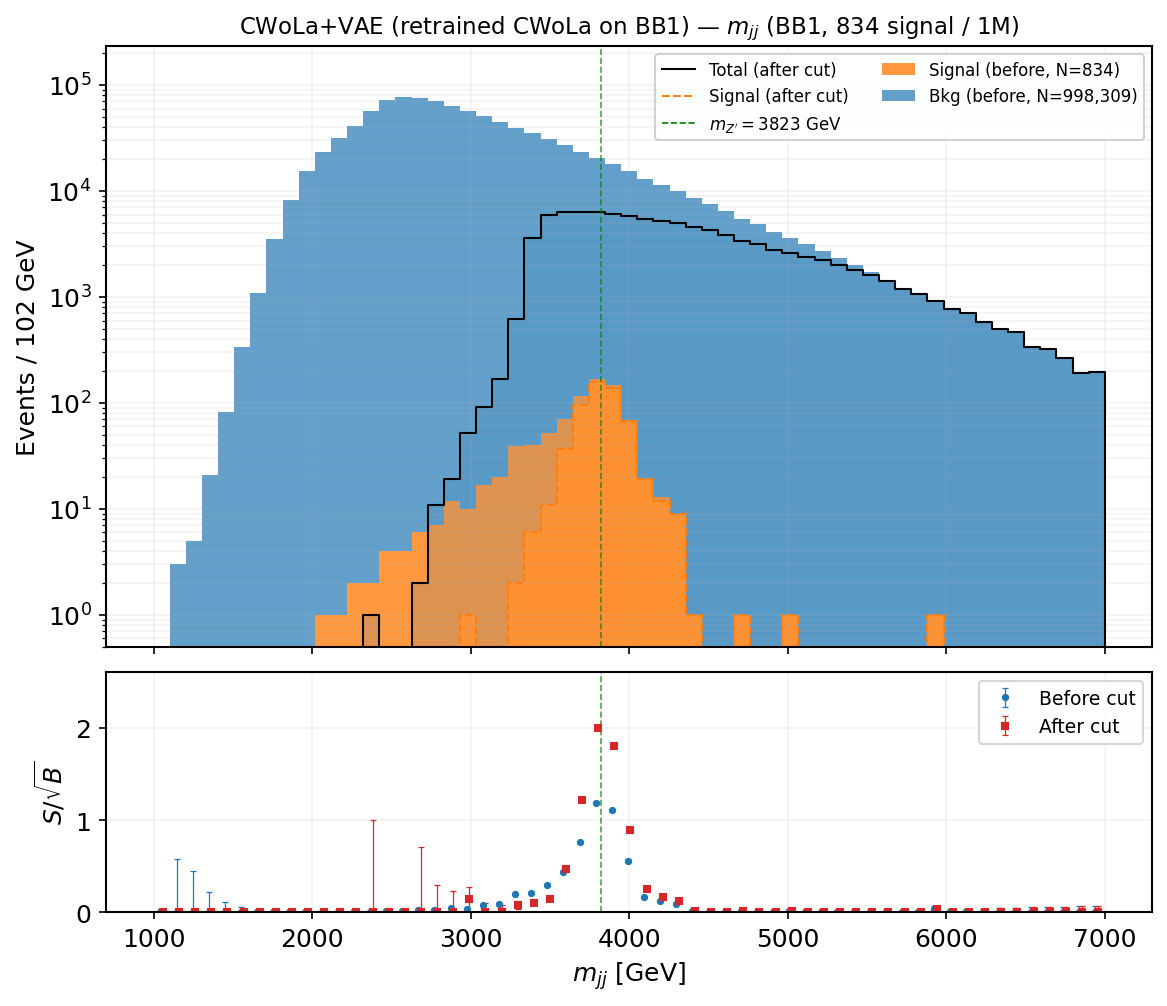}
    \caption{Distribution of the reconstructed dijet mass $m_{\text{JJ}}$ of the R\&D dataset (upper panel). The enhancement of the signal-over-background ratio (lower panel) with the cut of AD score reflects the performance of the AD algorithm. }
    \label{fig:mJJ}
\end{figure}

The three physics observables $m_{\text{J}_1}$ (Fig.~\ref{fig:mJ1}), $m_{\text{J}_2}$ (Fig.~\ref{fig:mJ2}), and $m_\text{JJ}$ (Fig.~\ref{fig:mJJ}) are designed to reconstruct two heavy BSM particles, \(X\) and \(Y\), produced in the decay of a third heavy BSM particle, \(Z'\). 
Both \(X\) and \(Y\) are assumed to decay into two quarks and are reconstructed as the two large-radius jets, \(J_1\) and \(J_2\). The reconstructed jet masses, \(m_{J_1}\) and \(m_{J_2}\) reflect the invariant masses of the parent particles \(X\) and \(Y\). The \(Z'\) candidate is reconstructed from the dijet system formed by \(J_1\) and \(J_2\), so its mass is captured by the corresponding \(m_{JJ}\) observable. These variables are therefore sensitive to the resonant structure of the BSM signal and provide discrimination from background processes.
However, because of the enormous background, the original signal-to-background ratio, or equivalently the significance defined as \(S/\sqrt{B}\), is poor and insufficient to establish the presence of a signal. 

The anomaly detection (AD) method, which aims to suppress background while retaining as much model-unspecific signal as possible, can therefore enhance the signal significance. In the plots, a clear enhancement of the resonance peaks is observed after applying the AD method, implemented by a selection of events identified as anomalous based on the output model score. The significance shown in the bottom panel exhibits an improvement of up to a factor of two, indicating the potential of this method for application to real data in searches for new physics, should such signals exist.
We also observe slight mass "sculpting", a common phenomenon in HEP analysis where correlations introduced by selection methods can distort key observables away from physical meaningfulness. 
This is not realized by the agentic workflow alone and highlights the need to incorporate expert-level domain-specific knowledge.
Limited by scope, we do not pursue a detailed analysis and optimization here, but present them for future reference. The performance and time cost of the three runs of this challenge task are summarized in Tab.~\ref{tab:lhco_summary}.

\begin{table}[htbp]
\centering
\small
\begin{tabular}{l c c c c}
\toprule
\textbf{Task Desc.} & \textbf{Iters} & \textbf{Wall time} & \textbf{R\&D AUC (signal 9.1\%)} & \textbf{BB1 AUC (signal 0.08\%)} \\
\midrule
Open-ended & 114 & $\sim$3.7h & 0.501 (fail) & - \\
Interactive & 160 & $\sim$9h & 0.830 & 0.886 \\
Guided & 46 & $\sim$1.5h & 0.854 & 0.885 \\
\bottomrule
\end{tabular}
\caption{Performance comparison of different variants of agentic approach to the LHCO AD challenge.}
\label{tab:lhco_summary}
\end{table}

Experiment 4 illustrates that at this stage, agentic-assisted workflow can provide a well-defined and correct direction for open-ended tasks.
However, human expertise and domain specific knowledge such as awareness of mass sculpting in the LHCO AD task are still necessary. In addition, the path  explored by the agentic system still remains largely a straightforward combination of existing research, and its capability for genuinely creative exploration is still uncertain. Human interaction, as a fundamental component of scientific research, we always believe, remains essential and cannot be fully replaced.

We argue further that this work demonstrates the successful combination of a carefully designed agentic system with interaction from external intelligence and suggests a clear scaling roadmap for the future. 
With continued development of backbone LLM models and appropriately gauged human-in-the-loop interventions, the system has the potential to evolve from solving closed-loop tasks to semi-open application and, eventually, to fully open-ended challenges. 
With domain-specific post-training of the LLM and with stronger reasoning and reflection capabilities, it is envisioned to eventually enable fundamentally new possibilities for scientific discovery.

\section{Conclusions}

We have presented the SciFi agentic AI framework for scientific workflows characterized by closed-loop task definitions and flexible, domain-specific implementations. By enclosing the agent system within a container-based execution environment, the framework supports safe and fully autonomous operation without routine confirmation at each step, thereby enabling the offloading of well-defined tasks in scientific research. 
Its minimalist, LLM-native design makes the system easier to review, maintain, and extend, while also providing a user-friendly natural-language interface for high-level interaction.

To address the requirements of scientific applications, the framework incorporates self-assessing model and workflow-review agents that emphasize verification, robustness, and the reduction of false positives. The framework has been evaluated on diverse tasks in the high energy physics domain of varying complexity.
These experiments use a wide range of task description levels and model backbones, including frontier proprietary models, cost-efficient commercial models, and fully open-source solutions. 
The results show that the workflow can operate effectively across models with different capability, cost, and openness characteristics, providing a practical path for routine closed-loop scientific tasks.

Overall, this study serves as a starting point for exploring the well-scoped use of agentic AI with apt human-in-the-loop intervention in scientific research. By focusing on closed-loop, well-defined tasks, the proposed framework aims to shift human effort away from repetitive technical execution and toward higher-level activities, including creative technology development and open-ended scientific inquiry.
These results have exciting and indeterminable potential impact on the scientific enterprise, with the prospect to usher in a new era of scientific research and discovery.

\ack{This work is supported by the U.S. Department of Energy under contract number DE-AC02-76SF00515.}

\appendix

\section{Appendix: Task Descriptions for Firmware and LHCO Challenge Experiments}
\label{app:task}

\begin{figure}[htbp]
\centering 

\begin{tcolorbox}[enhanced, sharp corners, boxrule=0.5pt,
    colframe=framegray, colback=white, width=6.6in,
    left=8pt, right=8pt, top=6pt, bottom=6pt,
    title={\sffamily\bfseries Firmware From Scratch  --- Task Description (NL) and Result},
    coltitle=white, colbacktitle=titlebg, fonttitle=\normalsize]

\tsec{Context}
{\scriptsize\sffamily
FPGA streaming inference project under \texttt{sim/}. Both the Verilog wrapper
(\texttt{sim/src/stream\_wrapper.v}) and the C++ binder (\texttt{sim/stream\_wrapper\_binder.cc})
are empty stubs --- design and write both from scratch. The conv kernel and dense layer
submodules in \texttt{sim/src/} are complete and must not be modified.
\texttt{sim/verify\_golden.py} builds the code via \texttt{make}, loads the resulting \texttt{.so},
calls an \texttt{inference} function through ctypes, and checks outputs against a golden dataset.

\smallskip
The design is a streaming pipeline: the binder feeds 25 ten-bit values per clock into a conv kernel
(\texttt{kernel\_wrapper}, 250-bit input, 90-bit output), you accumulate 20 kernel outputs in a
shift register (1800 bits), then a dense layer (\texttt{dense\_wrapper}) reads the full register and
produces a 14-bit output. The binder must handle random input pauses and capture outputs when
\texttt{out\_valid} pulses. Toolchain (verilator, g++, make, python3) is on PATH.
}

\rowsep
\tsec{Todo}
{\scriptsize\sffamily
\begin{enumerate}[nosep,leftmargin=1.5em,topsep=0pt]
  \item Read the existing submodules, binder contract, verify script, and build system.
  \item Design and write both \texttt{stream\_wrapper.v} and \texttt{stream\_wrapper\_binder.cc}.
  \item Build and run both verify modes (\texttt{-{}-no-pause} and \texttt{-{}-inp-pause 0.3 -{}-seed 42}).
\end{enumerate}
}

\rowsep
\tsec{Expect}
{\scriptsize\sffamily
\begin{itemize}[nosep,leftmargin=1em,topsep=0pt]
  \item Both verify modes print \texttt{PASSED: All} against the golden dataset
  \item Both files implemented (no longer empty stubs); no other \texttt{sim/src/} files modified
\end{itemize}
}

\rowsep

\tsec{Output --- two empty stubs $\to$ complete implementation}

\vspace{4pt}

\begin{minipage}[t]{0.47\linewidth}
\centering{\scriptsize\sffamily\bfseries\color{labelv} stream\_wrapper.v (agent)}

\vspace{2pt}
{\scriptsize\ttfamily\color{codefg}
\begin{tabular}[t]{@{}l@{}}
\colorbox{addbg}{\textcolor{addfg}{module stream\_wrapper(}} \\
\colorbox{addbg}{\textcolor{addfg}{~~clk, rst\_n, inp\_valid,}} \\
\colorbox{addbg}{\textcolor{addfg}{~~inp\_data, out\_valid, out\_data);}} \\
\colorbox{addbg}{\textcolor{addfg}{wire [89:0] kernel\_out;}} \\
\colorbox{addbg}{\textcolor{addfg}{kernel\_wrapper kernel\_inst(}} \\
\colorbox{addbg}{\textcolor{addfg}{~~.model\_inp(inp\_data),}} \\
\colorbox{addbg}{\textcolor{addfg}{~~.model\_out(kernel\_out));}} \\
\colorbox{addbg}{\textcolor{addfg}{reg [4:0] count;}} \\
\colorbox{addbg}{\textcolor{addfg}{reg [1799:0] shift\_reg;}} \\
\colorbox{addbg}{\textcolor{addfg}{always @(posedge clk or negedge rst\_n)}} \\
\colorbox{addbg}{\textcolor{addfg}{~~if (!rst\_n) begin}} \\
\colorbox{addbg}{\textcolor{addfg}{~~~~count<=0; shift\_reg<=0;}} \\
\colorbox{addbg}{\textcolor{addfg}{~~end else if (inp\_valid) begin}} \\
\colorbox{addbg}{\textcolor{addfg}{~~~~shift\_reg <= \{kernel\_out,}} \\
\colorbox{addbg}{\textcolor{addfg}{~~~~~~shift\_reg[1799:90]\};}} \\
\colorbox{addbg}{\textcolor{addfg}{~~~~if (count==19) begin}} \\
\colorbox{addbg}{\textcolor{addfg}{~~~~~~count<=0; out\_valid<=1;}} \\
\colorbox{addbg}{\textcolor{addfg}{~~~~end else begin}} \\
\colorbox{addbg}{\textcolor{addfg}{~~~~~~count<=count+1; out\_valid<=0;}} \\
\colorbox{addbg}{\textcolor{addfg}{~~end end}} \\
\colorbox{addbg}{\textcolor{addfg}{dense\_wrapper dense\_inst(}} \\
\colorbox{addbg}{\textcolor{addfg}{~~.model\_inp(shift\_reg),}} \\
\colorbox{addbg}{\textcolor{addfg}{~~.model\_out(out\_data));}} \\
\end{tabular}
}
\end{minipage}%
\hfill
\begin{minipage}[t]{0.50\linewidth}
\centering{\scriptsize\sffamily\bfseries\color{labelcc} stream\_wrapper\_binder.cc (agent)}

\vspace{2pt}
{\scriptsize\ttfamily\color{codefg}
\begin{tabular}[t]{@{}l@{}}
\colorbox{addbg}{\textcolor{addfg}{extern "C" size\_t inference(}} \\
\colorbox{addbg}{\textcolor{addfg}{~~const int32\_t *c\_inp, ...)\{}} \\
\colorbox{addbg}{\textcolor{addfg}{Vstream\_wrapper* top = new ...;}} \\
\colorbox{addbg}{\textcolor{addfg}{std::mt19937 gen(seed);}} \\
\colorbox{addbg}{\textcolor{addfg}{top->rst\_n=0; top->clk=0;}} \\
\colorbox{addbg}{\textcolor{addfg}{top->eval(); {\color{skipfg}// reset cycle}}} \\
\colorbox{addbg}{\textcolor{addfg}{top->clk=1; top->eval();}} \\
\colorbox{addbg}{\textcolor{addfg}{top->rst\_n=1; top->eval();}} \\
\\
{\color{skipfg}// for each sample:} \\
\colorbox{addbg}{\textcolor{addfg}{for (size\_t s=0; s<n\_samples; s++) \{}} \\
\colorbox{addbg}{\textcolor{addfg}{~~int i=0;}} \\
\colorbox{addbg}{\textcolor{addfg}{~~while (i<20) \{}} \\
\colorbox{addbg}{\textcolor{addfg}{~~~~if (dist(gen)>=inp\_pause\_prob) \{}} \\
\colorbox{addbg}{\textcolor{addfg}{~~~~~~top->inp\_valid = 1;}} \\
\colorbox{addbg}{\textcolor{addfg}{~~~~~~write\_input<25,10>(}} \\
\colorbox{addbg}{\textcolor{addfg}{~~~~~~~~top->inp\_data, ptr+i*25);}} \\
\colorbox{addbg}{\textcolor{addfg}{~~~~~~i++;}} \\
\colorbox{addbg}{\textcolor{addfg}{~~~~\} else top->inp\_valid = 0;}} \\
\colorbox{addbg}{\textcolor{addfg}{~~~~top->clk=1; top->eval();}} \\
\colorbox{addbg}{\textcolor{addfg}{~~~~if (top->out\_valid)}} \\
\colorbox{addbg}{\textcolor{addfg}{~~~~~~c\_out[n++] = top->out\_data;}} \\
\colorbox{addbg}{\textcolor{addfg}{~~~~top->clk=0; top->eval();}} \\
\colorbox{addbg}{\textcolor{addfg}{\}\}}} \\
\end{tabular}
}
\end{minipage}

\end{tcolorbox}

\label{fig:fw_v_n} 
\end{figure}

\clearpage

\begin{figure}[htbp]
\centering 

\begin{tcolorbox}[enhanced, sharp corners, boxrule=0.5pt,
    colframe=framegray, colback=white, width=6.6in,
    left=8pt, right=8pt, top=6pt, bottom=6pt,
    title={\sffamily\bfseries Firmware From Scratch  --- Task Description (EX) and Result},
    coltitle=white, colbacktitle=titlebg, fonttitle=\normalsize]

\tsec{Context}
{\scriptsize\sffamily
Implementing a Verilator-based simulation harness for an FPGA streaming inference pipeline FROM SCRATCH.
Both halves are EMPTY: \texttt{stream\_wrapper.v} (module declaration only) and
\texttt{stream\_wrapper\_binder.cc} (comments only). You design BOTH sides of the C++/Verilog interface.
The only fixed boundary is the Python$\to$C++ \texttt{inference} function signature.

\smallskip
Available submodules (complete, do NOT modify): \texttt{kernel\_wrapper.v} (combinational, 250-bit$\to$90-bit),
\texttt{dense\_wrapper.v} (combinational, 1800-bit$\to$14-bit).
Support files: \texttt{build\_binder.mk}, \texttt{ioutil.hh} (bit-packing helpers), \texttt{verify\_golden.py} (oracle).

\smallskip
\textbf{Required dataflow:} 500 int32 values per waveform (10 bits each). Chunk size = 25 values/clock
$\to$ 20 clocks per waveform (II=20). Kernel: 250-bit input, 90-bit output. Shift register: 20$\times$90 = 1800 bits.
Dense: 1800-bit$\to$14-bit. Binder drives \texttt{inp\_valid} with random pauses (\texttt{inp\_pause\_prob}),
captures output on \texttt{out\_valid} pulse.

\smallskip
\textbf{Required C contract:}
\texttt{extern "C" size\_t inference(const int32\_t *c\_inp, int32\_t *c\_out, size\_t n\_samples, double inp\_pause\_prob, uint32\_t seed);}

\smallskip
\textbf{Verilator strictness} (\texttt{-Wall}): handle \texttt{PROCASSINIT}, \texttt{EOFNEWLINE},
\texttt{WIDTHEXPAND}/\texttt{WIDTHTRUNC}, \texttt{UNUSEDSIGNAL} lint pragmas.
}

\rowsep
\tsec{Todo}
{\scriptsize\sffamily
\begin{enumerate}[nosep,leftmargin=1.5em,topsep=0pt]
  \item Read every authoritative file: \texttt{kernel\_wrapper.v}, \texttt{dense\_wrapper.v}, \texttt{verify\_golden.py}, \texttt{ioutil.hh}, \texttt{build\_binder.mk}, empty stubs.
  \item Design the \texttt{.v} module: port list, shift register, counter, control logic, submodule instantiations.
  \item Write \texttt{sim/src/stream\_wrapper.v}.
  \item Design the C++ binder: Verilator headers, \texttt{extern "C" inference(...)}, cycle loop with random pauses.
  \item Write \texttt{sim/stream\_wrapper\_binder.cc}.
  \item Build: \texttt{make -f build\_binder.mk clean \&\& make -f build\_binder.mk slow}.
  \item Verify both modes (\texttt{-{}-no-pause} and \texttt{-{}-inp-pause 0.3 -{}-seed 42}). Write \texttt{notes.md}.
\end{enumerate}
}

\rowsep
\tsec{Expect}
{\scriptsize\sffamily
\begin{itemize}[nosep,leftmargin=1em,topsep=0pt]
  \item \texttt{nopause.log} and \texttt{paused.log} both contain \texttt{PASSED: All}
  \item \texttt{notes.md} exists with a one-paragraph summary
  \item Both files implemented; no other \texttt{sim/src/} files modified; support files unchanged
\end{itemize}
}

\rowsep

\tsec{Output --- two empty stubs $\to$ complete implementation}

\vspace{4pt}

\begin{minipage}[t]{0.47\linewidth}
\centering{\scriptsize\sffamily\bfseries\color{labelv} stream\_wrapper.v (agent)}

\vspace{2pt}
{\scriptsize\ttfamily\color{codefg}
\begin{tabular}[t]{@{}l@{}}
\colorbox{addbg}{\textcolor{addfg}{module stream\_wrapper(}} \\
\colorbox{addbg}{\textcolor{addfg}{~~clk, rst, inp\_valid,}} \\
\colorbox{addbg}{\textcolor{addfg}{~~inp\_data, out\_valid, out\_data);}} \\
\colorbox{addbg}{\textcolor{addfg}{wire [89:0] kernel\_out;}} \\
\colorbox{addbg}{\textcolor{addfg}{kernel\_wrapper kernel\_inst(}} \\
\colorbox{addbg}{\textcolor{addfg}{~~.model\_inp(inp\_data),}} \\
\colorbox{addbg}{\textcolor{addfg}{~~.model\_out(kernel\_out));}} \\
\colorbox{addbg}{\textcolor{addfg}{dense\_wrapper dense\_inst(}} \\
\colorbox{addbg}{\textcolor{addfg}{~~.model\_inp(dense\_inp\_pipe),}} \\
\colorbox{addbg}{\textcolor{addfg}{~~.model\_out(dense\_out));}} \\
\colorbox{addbg}{\textcolor{addfg}{assign out\_data = dense\_out;}} \\
\colorbox{addbg}{\textcolor{addfg}{always @(posedge clk)}} \\
\colorbox{addbg}{\textcolor{addfg}{~~if (rst) begin}} \\
\colorbox{addbg}{\textcolor{addfg}{~~~~dense\_inp\_pipe<=0; count<=0;}} \\
\colorbox{addbg}{\textcolor{addfg}{~~end else if (inp\_valid) begin}} \\
\colorbox{addbg}{\textcolor{addfg}{~~~~case (count)}} \\
\colorbox{addbg}{\textcolor{addfg}{~~~~~~5'd0: pipe[89:0]<=kernel\_out;}} \\
\colorbox{addbg}{\textcolor{addfg}{~~~~~~5'd1: pipe[179:90]<=kernel\_out;}} \\
\colorbox{addbg}{\textcolor{addfg}{~~~~~~...}} \\
\colorbox{addbg}{\textcolor{addfg}{~~~~~~5'd19: pipe[1799:1710]<=kernel\_out;}} \\
\colorbox{addbg}{\textcolor{addfg}{~~~~endcase}} \\
\colorbox{addbg}{\textcolor{addfg}{~~~~if (count==19) begin}} \\
\colorbox{addbg}{\textcolor{addfg}{~~~~~~out\_valid<=1; count<=0;}} \\
\colorbox{addbg}{\textcolor{addfg}{~~~~end else begin}} \\
\colorbox{addbg}{\textcolor{addfg}{~~~~~~out\_valid<=0; count<=count+1;}} \\
\colorbox{addbg}{\textcolor{addfg}{~~end end}} \\
\end{tabular}
}
\end{minipage}%
\hfill
\begin{minipage}[t]{0.50\linewidth}
\centering{\scriptsize\sffamily\bfseries\color{labelcc} stream\_wrapper\_binder.cc (agent)}

\vspace{2pt}
{\scriptsize\ttfamily\color{codefg}
\begin{tabular}[t]{@{}l@{}}
\colorbox{addbg}{\textcolor{addfg}{extern "C" size\_t inference(}} \\
\colorbox{addbg}{\textcolor{addfg}{~~const int32\_t *c\_inp, ...)\{}} \\
\colorbox{addbg}{\textcolor{addfg}{Vstream\_wrapper* top = new ...;}} \\
\colorbox{addbg}{\textcolor{addfg}{std::mt19937 gen(seed);}} \\
\colorbox{addbg}{\textcolor{addfg}{std::bernoulli\_distribution}} \\
\colorbox{addbg}{\textcolor{addfg}{~~pause\_dist(inp\_pause\_prob);}} \\
\\
\colorbox{addbg}{\textcolor{addfg}{while (sample\_idx < n\_samples) \{}} \\
\colorbox{addbg}{\textcolor{addfg}{~~top->rst=1; {\color{skipfg}// reset per sample}}} \\
\colorbox{addbg}{\textcolor{addfg}{~~top->clk=0; top->eval();}} \\
\colorbox{addbg}{\textcolor{addfg}{~~top->clk=1; top->eval();}} \\
\colorbox{addbg}{\textcolor{addfg}{~~top->rst=0; top->eval();}} \\
\colorbox{addbg}{\textcolor{addfg}{~~while (value\_idx < 500) \{}} \\
\colorbox{addbg}{\textcolor{addfg}{~~~~if (pause\_dist(gen))}} \\
\colorbox{addbg}{\textcolor{addfg}{~~~~~~top->inp\_valid = 0;}} \\
\colorbox{addbg}{\textcolor{addfg}{~~~~else \{}} \\
\colorbox{addbg}{\textcolor{addfg}{~~~~~~top->inp\_valid = 1;}} \\
\colorbox{addbg}{\textcolor{addfg}{~~~~~~write\_input<25,10>(}} \\
\colorbox{addbg}{\textcolor{addfg}{~~~~~~~~top->inp\_data, \&c\_inp[...]);}} \\
\colorbox{addbg}{\textcolor{addfg}{~~~~~~value\_idx += 25; \}}} \\
\colorbox{addbg}{\textcolor{addfg}{~~~~top->clk=1; top->eval();}} \\
\colorbox{addbg}{\textcolor{addfg}{~~~~if (top->out\_valid)}} \\
\colorbox{addbg}{\textcolor{addfg}{~~~~~~read\_output<1,14>(}} \\
\colorbox{addbg}{\textcolor{addfg}{~~~~~~~~top->out\_data, \&c\_out[n]);}} \\
\colorbox{addbg}{\textcolor{addfg}{~~~~top->clk=0; top->eval();}} \\
\colorbox{addbg}{\textcolor{addfg}{\}\}}} \\
\end{tabular}
}
\end{minipage}

\end{tcolorbox}

\label{fig:fw_v_e} 
\end{figure}

\begin{figure}[htbp]
\centering 
\includestandalone[width=0.75\textwidth]{task_lhco/task-desc-lhco-openended}
\label{fig:lhco_o} 
\end{figure}

\begin{figure}[htbp]
\centering 
\includestandalone[width=0.75\textwidth]{task_lhco/task-desc-lhco-iterative}
\label{fig:lhco_i} 
\end{figure}

\begin{figure}[htbp]
\centering 
\includestandalone[width=0.75\textwidth]{task_lhco/task-desc-lhco-guided}
\label{fig:lhco_g} 
\end{figure}

\clearpage

\bibliographystyle{unsrt}
\bibliography{ref}

\end{document}